\documentclass[10pt,journal]{IEEEtran} 
\usepackage[nocompress]{cite}
\usepackage[utf8]{inputenc}
\usepackage{xcolor}
\usepackage{graphicx}
\usepackage{subfigure}
\usepackage[normalem]{ulem}
\usepackage{array}
\usepackage{tikz}
\usetikzlibrary{positioning}
\usetikzlibrary{calc}
\usepackage{forest}
\usepackage{pgfplots}
\usepackage{pgfplotstable}
\pgfplotsset{compat=1.17}
\usepackage{colortbl}
\usepackage{booktabs}
\usepackage{url}
\usepackage[hidelinks]{hyperref}
\usepackage{amsmath}
\usepackage{amssymb}
\usepackage{placeins}
\usepackage{orcidlink}
\usepackage{multirow}

\usepackage{enumitem}

\title{Metrics for Dataset Demographic Bias: A Case Study on Facial Expression Recognition}
\author{Iris~Dominguez-Catena\orcidlink{0000-0002-6099-8701},~\IEEEmembership{Student Member,~IEEE,} 
Daniel~Paternain\orcidlink{0000-0002-5845-887X},~\IEEEmembership{Member,~IEEE,}
Mikel~Galar\orcidlink{0000-0003-2865-6549},~\IEEEmembership{Member,~IEEE}%
\IEEEcompsocitemizethanks{\IEEEcompsocthanksitem I.Dominguez-Catena, D. Paternain and M. Galar are with the Department of Statistics, Computer Science and Mathematics, Public University of Navarre (UPNA), Arrosadia Campus, 31006, Pamplona, Spain and with the Institute of Smart Cities (ISC), Public University of Navarre (UPNA), Arrosadia Campus, 31006, Pamplona, Spain.\protect\\
E-mail: iris.dominguez@unavarra.es; daniel.paternain@unavarra.es; mikel.galar@unavarra.es}

\thanks{This work was funded by a predoctoral fellowship from the Research Service of the Universidad Publica de Navarra, the Spanish MICIN (PID2019-108392GB-I00,  PID2020-118014RB-I00 and PID2022-136627NB-I00/AEI/10.13039/501100011033 FEDER, UE), and the Government of Navarre (0011-1411-2020-000079 - Emotional Films). Open access funding provided by Universidad Pública de Navarra.
}}

\begin{document}

\bstctlcite{IEEEexample:BSTcontrol}

\maketitle

\begin{abstract}
  Demographic biases in source datasets have been shown as one of the causes of unfairness and discrimination in the predictions of Machine Learning models. One of the most prominent types of demographic bias are statistical imbalances in the representation of demographic groups in the datasets. In this paper, we study the measurement of these biases by reviewing the existing metrics, including those that can be borrowed from other disciplines. We develop a taxonomy for the classification of these metrics, providing a practical guide for the selection of appropriate metrics. To illustrate the utility of our framework, and to further understand the practical characteristics of the metrics, we conduct a case study of 20 datasets used in Facial Emotion Recognition (FER), analyzing the biases present in them. Our experimental results show that many metrics are redundant and that a reduced subset of metrics may be sufficient to measure the amount of demographic bias. The paper provides valuable insights for researchers in AI and related fields to mitigate dataset bias and improve the fairness and accuracy of AI models. The code is available at \url{https://github.com/irisdominguez/dataset_bias_metrics}.
\end{abstract}


\section{Introduction}

General advancements in technology, compounded with the widespread adoption of personal computers of all sorts, have led to an ever increasing exposure of society and non-expert users to autonomous systems. This interaction has also led to an accelerated deployment speed of state-of-the-art systems. Complex systems, such as general-purpose language and image models, conversational chatbots, or automatic face recognition systems, to name a few, are now deployed within months of their creation directly into the hands of nonexpert users. These systems, by their very nature, are difficult to evaluate and test, raising safety concerns. As systems interact with users in new and unpredictable ways, how can we ensure that no harm of any type is done to the user?

This general question is answered through the field of AI ethics~\cite{Jobin2019}. This field, in turn, takes shape in several other aspects, focusing on issues such as the integration of robotics in society~\cite{Winfield2019}, issues of digital privacy~\cite{Stahl2018}, and many others. One particularly interesting concept is algorithmic fairness~\cite{Mitchell2021}, which focuses on how systems can replicate human biases, discriminating people based on protected characteristics such as sex, gender, race, or age. Even if the concept of algorithmic fairness is broad and multifaceted, this notion of unwanted bias as the unwanted patterns learned by the machine makes them easier to characterize. In turn, the characterization and measurement of fairness favors the methodological mitigation of unfair behavior in trained models.

Although the development of bias is a complex phenomenon, deep learning techniques are especially susceptible to bias in datasets~\cite{Ntoutsi2020}. These techniques learn patterns autonomously and can often get confused between correlated patterns. When certain demographic characteristics are correlated with the target class of a problem, it is possible for the models to incorporate and amplify that correlation. This ends up resulting in a biased and differentiated prediction for certain individuals and demographic groups.

To recognize and solve these issues, it is crucial to measure bias, both in the final models and in the datasets. Although different metrics have been proposed~\cite{Verma2018}, most of them focus only on the bias exhibited by the trained models. The measurement of bias in the source datasets has not received the same attention, although it enables the validation of new bias mitigation methods~\cite{Zhao2017}, the explanation of bias transference throughout the training process~\cite{Dominguez-Catena2022}, and the demographic description of the application environment where a dataset or model can be safely used~\cite{Mitchell2019}.

In this article, we explore metrics that can be used to measure demographic bias in datasets. Most previous works that have focused on this issue~\cite{Dulhanty2019,Buolamwini2018} study biases only from an intuitive notion, without using bias metrics. A few works~\cite{Zhao2017,Kim2019} have employed metrics specific to bias, although only considering a single metric and without taking the different types of demographic bias into account. This work aims to serve as a unifying framework for the few metrics that have been already used for this purpose and those that can be adapted from other fields with equivalent problems, such as population diversity metrics used in ecology. This wider variety of metrics allows us to cover specific types of biases that were previously unmeasured. Accordingly, we propose a taxonomy for dataset bias metrics based on the type of bias measured, facilitating the selection of appropriate metrics. Based on this taxonomy, we aim to find a concrete, expressive, and interpretable set of metrics to facilitate the work of analyzing the demographic and bias properties of ML datasets. To our knowledge, no such taxonomy or metric selection has been previously proposed.

An interesting case study to evaluate our taxonomy and set of metrics is Facial Expression Recognition (FER). FER is a problem in which photographs of people are classified according to the emotion they appear to express. The applications of this problem are varied, including healthcare~\cite{Werner2022} and assistive robotics~\cite{Nimmagadda2022} among others, and in most cases involve direct interaction with non-expert users. Unlike other problems in which bias is usually studied, mainly based on tabular data and with explicit demographic information~\cite{Berk2018}, in FER the demographics of the person are rarely known. However, these demographic factors directly influence the person's appearance, generating demographic proxies that are completely embedded in the input images, making FER an interesting problem from the AI bias perspective. Even if this information is masked, it has been shown that both emotion expression and emotion identification are conditioned by a person's demographics~\cite{Jack2009,Elfenbein2002}, so differential treatment based on demographics may be unavoidable.

More specifically, we gather twenty FER datasets according to a clear set of criteria, obtain a demographic profile of each of them, and apply the reviewed dataset bias metrics. We then employ these results to explore both the characteristics and limitations of each metric. We use this information to select a set of non-redundant interpretable metrics that can summarize the demographic biases in a dataset. Additionally, we employ the metrics to assess the types of biases found in FER datasets, where we observe differences between datasets created from different data sources. Identifying the different bias profiles of specific datasets can improve both the choice of training datasets and the choice of mitigation techniques.

Although this paper focuses on dataset bias, in the Supplementary Material we also provide the results of a series of experiments on the downstream propagation of such biases to the trained model, showing the importance of the appropriate characterization of the different types of dataset demographic bias.


The following sections are as follows. First, Section~\ref{sec:related} recalls some related work in the field. Subsequently, in Section~\ref{sec:metrics} we review and present a taxonomy of dataset demographic bias metrics. Section~\ref{sec:experiments} then presents the experimental framework for the FER case study, while Section~\ref{sec:results} gathers the results found in these experiments. Finally, Section~\ref{sec:conclusion} summarizes our findings and potential future work.

\section{Related work}\label{sec:related}

This section introduces some relevant background for our work. In Section~\ref{ssec:fairness}, we provide a brief overview of fairness, its relationship to bias, and the methods used to measure it. Subsequently, in Section~\ref{ssec:fer}, we explore the application of these concepts in the context of FER.

\subsection{Fairness}\label{ssec:fairness}

The advances in ML and Artificial Intelligence (AI) in the last decades have led to an explosion of real-world applications involving intelligent agents and systems. This has inevitably led researchers to consider the social implications of these technologies and study what fairness means in this context~\cite{Danks2017, Verma2018, Barocas2019, Mehrabi2021, Schwartz2022}. The efforts to develop technical standards and best practices have also generated an increasing number of guidelines for ethical AI and algorithmic fairness~\cite{Jobin2019}.

Most definitions of fairness revolve around the concept of unwanted bias~\cite{Mehrabi2021}, also known as prejudice or favoritism, where particular individuals or groups defined by certain protected attributes, such as age, race, sex, and gender, receive an undesired differentiated treatment. In this sense, an algorithm or system is defined as fair if it is free of unwanted bias. It is important to note that although the concept of demographic bias is related to bias in machine learning and many results can be adapted to both~\cite{Torralba2011}, the particularities and potential harm resulting from demographic bias require an independent study. To this end, different metrics and mathematical definitions have been designed to characterize both fairness and demographic bias~\cite{Verma2018,Das2021,Mehrabi2021}. It is important to note that together with these metrics and definitions, criticism has also arisen~\cite{Schwartz2022,Thomas2020}, since an excessive optimization of any given quantitative metric can lead to a loss of meaning, resulting in a false impression of fairness. As the fairness definitions and metrics proposed in the literature~\cite{Mehrabi2021,Das2021} are mostly concerned with the social impact of the deployed systems, they deal only with the presence of bias in the final trained model, regardless of the source of that bias. These definitions, such as \textit{equalized odds}~\cite{Hardt2016}, \textit{equal opportunity}~\cite{Hardt2016} or \textit{demographic parity}~\cite{Zafar2017}, are usually designed to detect a disparate treatment between a single \textit{priviledged} demographic group and a single \textit{protected} demographic group, in classification problems where one of the classes is considered preferable (usually the \textit{positive} class). Despite this classical perspective, recent works~\cite{Suresh2021,Mehrabi2021,Schwartz2022} have focused on the multiple complementary sources that can lead to unwanted bias in the final model. These sources of bias originate in different phases of the AI pipeline~\cite{Suresh2021}, such as data collection, model training, and model evaluation. Regarding practical applications, these definitions and taxonomies of bias have been applied to multiple domains, where demographic biases have been found in facial~\cite{Grother2019, Dooley2021} and gender~\cite{Keyes2018, Buolamwini2018} recognition, to name a few.

The bias detected in the source data has been a topic of particular interest over the years~\cite{Ntoutsi2020,Prabhu2020}. Large public datasets have become the basic entry point for many AI projects, where developers often use them with limited knowledge of the origin of the data and its biases~\cite{Prabhu2020}. Some of the most popular ML datasets, such as ImageNet~\cite{Deng2009, Denton2021} and COCO~\cite{Zhao2021}, have been revealed to include severe demographic biases~\cite{Dulhanty2019} and even direct examples of racism, such as racial slurs~\cite{Prabhu2020}. To identify such biases, some auxiliary datasets have been proposed, either by annotating previous datasets for apparent demographic characteristics~\cite{Zhao2021,Garcia2023}, or developing new demographically annotated datasets~\cite{Hazirbas2022,Porgali2023}, which can then be used to evaluate models and datasets. However, few works~\cite{Dulhanty2019,Buolamwini2018,Zhao2017,Kim2019,Wang2022b} have focused on the measurement and mathematical characterization of the bias present in the source data. Some works~\cite{Dulhanty2019,Buolamwini2018} focus on non-systematic approaches, calculating the proportion of different demographic subgroups and manually looking for imbalances. Other works~\cite{Zhao2017,Kim2019} employ metrics based on information theory and statistics, such as the Mutual Information, to quantify bias in the source datasets, usually as part of a bias mitigation methodology, or focus on specific types of dataset, such as object detection~\cite{Wang2022b}, where specific metrics relating to the nature of the problem can be developed.

To the best of our knowledge, no previous work has unified these approaches, systematically comparing the properties of the different metrics. In this work, we explore and classify the full array of metrics already in use to measure dataset demographic bias and propose the application of existing metrics from equivalent problems in other fields, especially in ecology.

\subsection{Facial Expression Recognition}\label{ssec:fer}

FER is the problem of automatic emotion recognition based on facial images. Although several variants exist, the most common implementation employs static images to identify a specific set of possible emotions. These range from smile~\cite{Chen2017} or pain~\cite{Werner2022} recognition, to the most widely used emotion classification proposed by Ekman~\cite{Ekman1971} (angry, disgust, fear, sad, surprise, and happy), with most publicly available datasets labeled with this codification. Although research has raised some concerns about the universality of both the underlying emotions~\cite{Ekman2016} and their associated facial expressions~\cite{Elfenbein2002, Chen2018}, the simplicity of the discrete codification and its labeling make it the most popular. 

Recent developments in Deep Learning (DL) and deep convolutional networks~\cite{LeCun2015}, technical advances such as the training of DL models on GPUs, together with the surge of larger datasets, have allowed the end-to-end treatment of FER as a simple classification problem, trained with supervised techniques based on labeled datasets. For this reason, numerous facial expression datasets have been published to aid in the development of FER. Several reviews have focused on collecting and analyzing available datasets over the years~\cite{Li2020,Guerdelli2022}, but to our knowledge, none of them has reviewed their demographic properties and potential biases. The FER datasets available differ in many aspects, including the source of data (internet, media, artificial, or gathered in laboratory conditions), the image or video format and technologies (visible spectrum, infrared, ultraviolet and 3D), the type of expressions registered (micro- and macro-expression), the emotion codification (continuous, discrete, and their variants) and the elicitation method (acted, induced, or natural). Demographically speaking, some datasets openly focus on facial expressions of specific demographic groups, such as JAFFE~\cite{Lyons1998, Lyons2021} (Japanese women) and iSAFE~\cite{Singh2020} (Indian people), but for the most part the datasets have not been collected taking diversity into account.

Some recent works have already found specific biases around gender, race, and age in both commercial FER systems~\cite{Kim2021, Ahmad2022} and research models~\cite{Greene2020,Xu2020,Domnich2021,Jannat2021,Deuschel2021,Poyiadzi2021}. From these works, Kim et al.~\cite{Kim2021} focus on the age bias of commercial models in an age-labeled dataset. Ahmad et al.~\cite{Ahmad2022} also study commercial models, but extend the research to age, gender, and race (considering two racial categories) by employing a custom database of politician videos. Regarding research models, Xu et al.~\cite{Xu2020} study age, gender, and race bias in models trained on an Internet search-gathered dataset and evaluated on a different dataset with known demographic characteristics. Two works~\cite{Domnich2021,Jannat2021} focus on gender bias in trained models. Deuschel et al.~\cite{Deuschel2021} analyzes biases in the prediction of action units with respect to gender and skin color in two popular datasets. Poyiadzy et al.~\cite{Poyiadzi2021} work on age bias in a dataset collected from Internet searches, performing additional work to generate apparent age labels for it. Additionally, some works~\cite{Xu2020, Jannat2021, Poyiadzi2021} have explored mitigation strategies applicable to this problem. According to the flaws and biases found in previous studies, Hernandez et al.~\cite{Hernandez2021} propose a set of guidelines to assess and minimize potential risks in the application of FER-based technologies.

Notwithstanding these previous works on specific biases and the resulting general guidelines, no other work has comparatively analyzed the demographic bias of a large selection of FER datasets or used multiple metrics to account for representational and stereotypical bias. We hope that the work presented here motivates new approaches to bias detection and mitigation in FER datasets.

\section{Dataset bias metrics}\label{sec:metrics}

Many specific metrics have been proposed to quantify bias and fairness~\cite{Pessach2020}. Unfortunately, most of these metrics only consider the disparity in the treatment of demographic groups in the trained model predictions. This type of bias, directly related to discrimination in the legal sense, disregards the source of the disparity. Additionally, the few metrics that have focused on dataset bias~\cite{Das2021} are only defined for binary classification problems, making the measurement of demographic biases in the source dataset an unexplored problem for more general multiclass classification problems.

In this section, our aim is to fill this gap by collecting metrics that are applicable to demographic bias in datasets, especially in multiclass problems. For this, we include both a few metrics that have previously been used in this context and, for the most part, metrics from other disciplines and contexts that can be adapted for this purpose, such as metrics from information theory (such as metrics based on \textit{Shannon entropy}) and ecology (such as \textit{Effective number of species}). 

\subsection{Taxonomy of demographic bias metrics}\label{ssec:taxonomy}

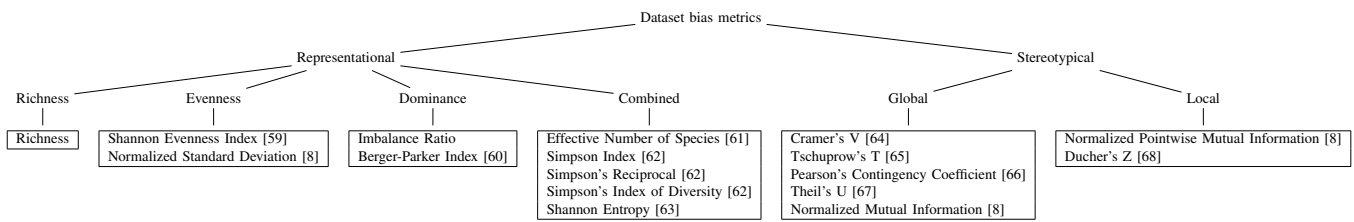
\begin{figure*}[hbt]
    \centering
    \resizebox{\textwidth}{!}{
    \begin{forest}
    [Dataset bias metrics
        [Representational
            [Richness   
                [\begin{tabular}{|l|}
                    \hline
                    Richness \\
                    \hline
                \end{tabular}]
            ]
            [Evenness
                [\begin{tabular}{|l|}
                    \hline
                    Shannon Evenness Index~\cite{Pielou1966} \\
                    Normalized Standard Deviation~\cite{Dominguez-Catena2022} \\
                    \hline
                \end{tabular}]
            ]
            [Dominance
                [\begin{tabular}{|l|}
                    \hline
                    Imbalance Ratio \\
                    Berger-Parker Index~\cite{Berger1970} \\
                    \hline
                \end{tabular}]
            ]
            [Combined
                [\begin{tabular}{|l|}
                    \hline
                    Effective Number of Species~\cite{Jost2006} \\
                    Simpson Index~\cite{Simpson1949} \\
                    Simpson’s Reciprocal~\cite{Simpson1949} \\
                    Simpson’s Index of Diversity~\cite{Simpson1949} \\
                    Shannon Entropy~\cite{Shannon1949}  \\
                    \hline
                \end{tabular}]
            ]
        ]
        [Stereotypical
            [Global
                [\begin{tabular}{|l|}
                    \hline
                    Cramer's V~\cite{Cramer1991} \\
                    Tschuprow's T~\cite{Tschuprow1939} \\
                    Pearson's Contingency Coefficient~\cite{Sakoda1977} \\
                    Theil's U~\cite{Theil1970} \\
                    Normalized Mutual Information~\cite{Dominguez-Catena2022} \\
                    \hline
                \end{tabular}]
            ]
            [Local
                [\begin{tabular}{|l|}
                    \hline
                    Normalized Pointwise Mutual Information~\cite{Dominguez-Catena2022} \\
                    Ducher's Z~\cite{Ducher1994} \\
                    \hline
                \end{tabular}]
            ]
        ]
    ]
\end{forest}}
    \caption{Taxonomy of dataset demographic bias metrics.}
    \label{fig:types_of_bias}
\end{figure*}

We propose a taxonomy of demographic bias metrics based on the two main types of statistical demographical bias, that is, representational and stereotypical bias. This coarse classification is then further refined into the families of metrics that can measure each bias. The proposed taxonomy is outlined with the associated metrics in Figure~\ref{fig:types_of_bias}. The following families of metrics can be identified. 

\vspace{0.3em}\noindent\textbf{Representational bias}. Representational bias refers to a lack of general demographic diversity in the dataset, that is, an unequal representation of demographic groups, such as having more samples from male presenting people than female presenting ones. This type of bias is not related to the target variable and, therefore, can be applied to any dataset, not only to classification tasks. A similar concept can be found in the field of ecology, where diversity, and more specifically the local-scale diversity of species in an ecosystem ($\alpha$-diversity~\cite{Whittaker1960}), is used as the opposite of representational bias. As the two concepts are directly related, one being the opposite of the other, any metric designed for diversity can be borrowed as a representational bias metric. Furthermore, in ecology diversity (and consequently representational bias) can be refined into three related components, namely \textit{Richness}, \textit{Evenness} and \textit{Dominance}. Several metrics target specific components, while others measure multiple components at the same time. We classify the latter as \textit{Combined} metrics. 

\begin{itemize}
    \item \textbf{Richness}. This category has a single metric, since it refers to the raw number of groups represented and constitutes the most direct and simple metric of representational bias. A dataset that has a representation of only a small number of groups (low richness) will be biased in favor of those groups, and susceptible to differentiated treatment in the form of inaccurate predictions in the trained model towards any group not represented. An example of a bias from lack of richness in a FER dataset would be having a dataset with only one racial group represented. 
    \item \textbf{Evenness}. Unfortunately, even if a dataset represents many groups, this representation may not be homogeneous, having a global overrepresentation of certain groups and an underrepresentation of others. The homogeneity of the group representation is also known as \textit{evenness}. For example, a FER dataset representationally biased by a lack of evenness would be one that, having representation of white, black and indian people, has an uneven composition of $45\%$ each of white and black people and only $10\%$ of Indian people.
    \item \textbf{Dominance}. Dominance refers to the population quota of the largest or \textit{dominant} group in the dataset. Dominance is not independent from Richness and Evenness, but it is a robust and easy to interpret notion, making its metrics common choices for representational bias measurement. Unfortunately, in exchange, it loses a lot of information related to the rest of the groups present, making them insufficient metrics when employed alone. For example, a FER dataset could be dominated by an age group if $95\%$ of the population is in the range $20-25$, with the remaining $5\%$ shared among the rest of the age groups. 
    \item \textbf{Combined}. Several metrics are not directly related to any of the other components and instead measure combinations of them. These metrics can usually summarize the amount of representational bias in a dataset, at the cost of not being able to distinguish the specific components of that bias.
\end{itemize}

\vspace{0.3em}\noindent\textbf{Stereotypical bias}. This type of bias can be identified when working on any labeled dataset, as an association between sample variables. These variables are usually a demographic property and the target variable in classification and regression tasks, although it can also be applied to spurious associations between several demographic properties. In stereotypical bias, the under- or overrepresentation is not directly found in a global view of the dataset, but in the specific demographic composition of each of the target classes. In FER, for example, a common stereotypical bias is an overrepresentation of female presenting people in the happy class~\cite{Dominguez-Catena2022}. Stereotypical bias can be measured at two levels:

\begin{itemize}
    \item \textbf{Global}. Global stereotypical bias refers to the global grade of association between a demographic component and the target classes, as a single measure of the whole dataset. For example, in FER, how related is the gender of the subjects and the target facial expressions.
    \item \textbf{Local}. Local stereotypical bias refers to how specific combinations of the demographic group and the target class are over- or underrepresented in the dataset. For example, in FER, if the proportion of female presenting subjects in the happy class in particular is above or below the expected proportion.
\end{itemize}

\subsection{Considerations for the metrics}

In this work, we consider classification problems where both the demographic groups of interest and the target classes are given as nominal variables. In regression problems, where the target variable is ordinal or numerical, or when one of the demographic groups can be codified as an ordinal or numerical variable, other metrics can provide more accurate information. Despite this, ordinal and numerical variables can be reinterpreted as nominal, making it a useful first approach. A good example is found in our case study, where the age variable is codified into age groups that can be treated as a nominal variable, although at the cost of some information loss. Regarding the target variable, although we focus on classification problems with a nominal target variable, the presented bias metrics can also be applied in other problems. Representational bias metrics do not consider the target variable, and as such can be directly applied in unlabeled datasets. Stereotypical bias metrics can also be applied in the same datasets, although in these cases they can only be used to measure the correlation between several demographic variables, rather than a demographic variable and the target variable, as is presented here.

An important consideration is the application of sampling corrections to the metrics. Many of the metrics, such as those from the field of ecology, are intended to be applied to a discrete sample of a larger population and subsequently corrected for sample size. In the case of AI datasets, we can either understand them as a sample from a global population and keep these corrections, or as complete populations and employ the uncorrected formulas. In practice, as models are mostly trained on a single dataset or a small set of them, we care more about the properties of the specific dataset than about their relationship to the real world population from which they were obtained. Thus, we do not employ the sample size correction variations, as they can hide and lower the bias of the smaller datasets. In certain techniques based on variable size datasets, such as the use of generative adversarial networks to generate datasets on demand~\cite{Kammoun2022}, sampling corrections must be applied when evaluating biases. We consider this case outside of the scope of this work and focus on fixed-size datasets.

Taking these considerations into account, the reviewed metrics and their properties are summarized in Table~\ref{table:metrics}. In the table, each row includes the information corresponding to a metric, namely, the full name and references, the symbol to be used in the rest of this work, the type and subtype of the metric according to the taxonomy presented in Section \ref{ssec:taxonomy}, the upper and lower bounds of the metric, and whether it directly (or inversely) measures bias. The metrics are further discussed in the following sections. As the metrics come from various sources, we present a unified mathematical formulation. The following unifying notation is employed:

\begin{itemize}
    \item We define $G$ as the set of demographic groups defined by the value of the protected attribute. For example, if $G$ stands for \textit{gender presentation}, a possible set of groups would be $\{\text{masculine}, \text{feminine}, \text{androgynous}\}$.
    \item We define $Y$ as the set of classes of the problem.
    \item We define $X$ as a population of $n$ samples.
    \item We define $n_a$ as the number of samples from the population $X$ that have the property $a$. For example, $n_g$ with $g \in G$, represents the number of samples in $X$ that corresponds to subjects belonging to the demographic group $g$. $a$ will usually be a demographic group $g \in G$, a target class $y \in Y$, or the combination of both properties $g\land y, g\in G, y \in Y$.
    \item Similarly, we define $p_a$ as the proportion of samples from the population $X$ that have the property $a$: $$p_a = \frac{n_a}{n}\;.$$
\end{itemize}

\begin{table*}[hbt]
  \begin{center}
    \caption{Summary of dataset demographic bias metrics and their characteristics.}
    \label{table:metrics}
    \pgfkeys{/pgf/number format/fixed}
    \begin{filecontents*}{data/metricas.csv}
"Name";"Symbol";"Type";"Subtype";"Range";"Relation to bias"
"Richness";"R";"representational";"richness";"$\left[0, \infty \right)$";"inverse"
"Shannon Evenness Index~\cite{Pielou1966}";"SEI";"representational";"evenness";"$\left(0, 1\right]$";"inverse"
"Normalized Standard Deviation~\cite{Dominguez-Catena2022}";"NSD";"representational";"evenness";"$\left(0, 1\right]$";"direct"
"Imbalance Ratio";"IR";"representational";"dominance";"$\left[1, \infty \right)$";"direct"
"Berger-Parker Index~\cite{Berger1970}";"BP";"representational";"dominance";"$\left[1/\text{R}, 1\right]$";"direct"
"Effective Number of Species~\cite{Jost2006}";"ENS";"representational";"combined";"$\left[1, \text{R}\right]$";"inverse"
"Simpson Index~\cite{Simpson1949}";"D";"representational";"combined";"$\left(0, 1\right]$";"direct"
"Simpson's Reciprocal~\cite{Simpson1949}";"1 / D";"representational";"combined";"$\left[1, \text{R}\right]$";"inverse"
"Simpson's Index of Diversity~\cite{Simpson1949}";"1-D";"representational";"combined";"$\left[0, 1\right)$";"inverse"
"Shannon Entropy~\cite{Shannon1949}";"H";"representational";"combined";"$\left[0, \ln\text{R}\right]$";"direct"
"Cramer's V~\cite{Cramer1991}";"$\phi_c$";"stereotypical";"global";"$\left[0, 1\right]$";"direct"
"Tschuprow's T~\cite{Tschuprow1939}";"T";"stereotypical";"global";"$\left[0, 1\right]$";"direct"
"Pearson's Contingency Coefficient~\cite{Sakoda1977}";"C";"stereotypical";"global";"$\left[0, 1\right]$";"direct"
"Theil's U~\cite{Theil1970}";"U";"stereotypical";"global";"$\left[0, 1\right]$";"direct"
"Normalized Mutual Information~\cite{Dominguez-Catena2022}";"NMI";"stereotypical";"global";"$\left[0, 1\right]$";"direct"
"Normalized Mutual Pointwise Information~\cite{Dominguez-Catena2022}";"NPMI";"stereotypical";"local";"$\left[-1, 1\right]$";"direct"
"Ducher's Z~\cite{Ducher1994}";"Z";"stereotypical";"local";"$\left[-1, 1\right]$";"direct"
\end{filecontents*}
\pgfplotstabletypeset[
        col sep=semicolon,
        text indicator=",
        columns={[index]0,
            [index]1,
            [index]2,
            [index]3,
            [index]4,
            [index]5
            },
        skip rows between index={17}{19}, 
        string type,
        display columns/0/.style={column type = {l}},
        display columns/1/.style={column type = {l}},
        display columns/2/.style={column type = {l}},
        display columns/3/.style={column type = {l}},
        display columns/4/.style={column type = {l}},
        display columns/5/.style={column type = {l}},
        every head row/.style={
            before row=\toprule,after row=\midrule},
        every last row/.style={
            after row=\bottomrule}
    ]{data/metricas.csv}
  \end{center}
\end{table*}

\subsection{Metrics for representational bias}\label{ssec:rep_metrics}

First, let us consider the \textbf{richness} metrics.

\vspace{0.3em}\noindent\textbf{Richness} (R)~\cite{Whittaker1960}. Richness is the simplest and most direct metric of diversity, which can be understood as the opposite of representational bias. Mathematically, richness is defined as:
\begin{equation}
    \text{R}(X) = |\{g\in G|n_g > 0\}|\;.
\end{equation}

Although Richness is a highly informative and interpretable metric, it disregards \textit{evenness} information on potential imbalances between group populations. In this sense, it can only assert that there are diverse examples through the dataset, but nothing about the proportions at which these examples are found. This metric is still vital for the interpretation of many other metrics, as they are either mathematically bounded by Richness or best interpreted when accompanied by it.

The following metrics focus on the \textbf{evenness} component of representational bias.

\vspace{0.3em}\noindent\textbf{Shannon Evenness Index} (SEI)~\cite{Pielou1966}. The most common example of \textit{evenness} metrics is the Shannon Evenness Index, a normalized version of the \textit{Shannon entropy} designed to be robust to \textit{Richness} variations. Due to this, only \textit{evenness}, the homogeneity of the groups present in the population, is taken into account. It is defined as:
\begin{equation}
    \text{SEI}(X) = \frac{\text{H}(X)} {\ln(\text{R}(X))}\;,
\end{equation}
where $\text{H}(X)$ is the Shannon entropy, defined later in Eq.~\ref{formula:shannon}.

As the entropy is divided by its theoretical maximum, corresponding to an even population of $\text{R}(X)$ groups, the metric is bounded between $0$ for uneven populations and $1$ for perfectly balanced ones. This value is independent for different number of represented groups, so that datasets with different R can achieve the same SEI.

\vspace{0.3em}\noindent\textbf{Normalized Standard Deviation} (NSD)~\cite{Dominguez-Catena2022}. Another metric for \textit{evenness} is the Normalized Standard Deviation of the population distribution. It is defined as:
\begin{equation}
    \text{NSD}(X) = \frac{|G|}{\sqrt{|G|-1}}\sqrt{\frac{\sum_{g \in G}(p_g - \bar{p})^2}{|G|}}\;,
\end{equation}
where $\overline{p}$ stands for the arithmetic mean of the population profile, which for a normalized profile is $\overline{p} = 1 / |G|$.

The normalization used in this metric produces the same upper and lower bounds as those of SEI. In this case, the metric is designed to target representational bias, so the meaning of the bounds is inverted, $0$ for the more balanced and even datasets, and $1$ for the extremely biased ones.

The third component of representational bias, \textbf{dominance}, is measured by the following metrics.

\vspace{0.3em}\noindent\textbf{Imbalance Ratio} (IR). The most common metric for class imbalance in AI is the \textit{Imbalance Ratio}. Although its most common application is measuring target class imbalance, the same metric can be used for any partitioning of a dataset, such as the one defined by a demographic component. It is defined as the population ratio between the most represented class and the least represented class:
\begin{equation}
    \text{IR}(X) = \frac{\displaystyle\max_{g\in G} n_g}{\displaystyle\min_{g \in G} n_g}\;.
\end{equation}

This definition leads to a metric that ranges from $1$ for more balanced populations to infinity for more biased ones. The inverse of this definition can be used to limit the metric between $0$ (exclusive) and $1$ (inclusive), with values close to $0$ indicating strongly biased populations and $1$ indicating unbiased ones. In this work, we employ this alternative $\text{IR}^{-1}(X)$ formulation.

This metric is commonly used for binary classification problems. When applied to more than two classes or groups, the metric simply ignores the rest of the classes, losing information in these cases.

\vspace{0.3em}\noindent\textbf{Berger-Parker Index} (BP)~\cite{Berger1970}. A metric closely related to the \textit{Imbalance Ratio} is the Berger-Parker Index. This metric measures the relative representation of the more abundant group relative to the whole population. As IR, it does not use all information of the population distribution, as imbalances between minority classes are not taken into account. It is defined as:
\begin{equation}
    \text{BP}(X) = \frac{\displaystyle\max_{g \in G} n_g}{n}\;.
\end{equation}

This metric is bounded between $1/\text{R}(X)$ and $1$, with values close to $1/\text{R}(X)$ indicating representationally unbiased datasets, and $1$ indicating biased ones.

Finally, some metrics measure representational bias as a \textbf{combination} of several components simultaneously.

\vspace{0.3em}\noindent\textbf{Effective Number of Species} (ENS)~\cite{Jost2006}. The Effective Number of Species is a robust measure that extends the \textit{Richness}, keeping the same bounds but integrating additional information about \textit{evenness}. This metric is upper bounded by $\text{R}(X)$, being equal to it for a totally balanced population, and smaller for increasingly biased populations, down to a lower bound of $1$ for populations with total dominance of a single group. It is defined as:
\begin{equation}
    \text{ENS}(X) = \exp\left({-\sum_{g\in G}{p_g \ln p_g}}\right)\;.
\end{equation}

The \textit{ENS} has several alternative formulations. The specific formula presented here is based on the \textit{Shannon entropy}. This means that this formulation is equivalent to the \textit{Shannon entropy} in their resulting ordering of populations. The difference lies only in the interpretability of the results, which are scaled to fit into the $[1, \text{R}(X)]$ range, with higher values indicating less representationally biased populations or datasets. The result is intended to follow the notion of an effective or equivalent number of equally represented groups. For example, a population with $\text{ENS}(X)=1.5$ is more diverse than one with one represented group ($\text{ENS}=1$) and less than one with two equally represented groups ($\text{ENS}=2$).

\vspace{0.3em}\noindent\textbf{Shannon entropy} (H)~\cite{Shannon1949}. The Shannon entropy, also known as \textit{Shannon Diversity Index} and \textit{Shannon-Wiener Index}, can also be directly used to measure diversity. In this case, diversity is measured by the amount of uncertainty, as defined by the entropy, with which we can predict to which group a random sample belongs. It is defined as:
\begin{equation}\label{formula:shannon}
    \text{H}(X) = -\sum_{g \in G} p_g \ln(p_g)\;.
\end{equation}

This metric lies in the range $[0,\ln\left(\text{R}(X)\right)]$, where $0$ identifies a population with a single represented group, and a value of $\ln\left(\text{R}(X)\right)$ corresponds to a perfectly balanced dataset composed of $\text{R}(X)$ different groups. 

\vspace{0.3em}\noindent\textbf{Simpson Index} (D)~\cite{Simpson1949}, \textbf{Simpson's Index of Diversity} ($1-\text{D}$), and \textbf{Simpson's Reciprocal} ($1 / \text{D}$). The Simpson Index is another metric influenced by both \textit{Richness} and \textit{evenness}. Mathematically, it is defined as:
\begin{equation}
    \text{D}(X) = \sum_{g \in G} p_g^2\;.
\end{equation}

This metric ranges from $1$ for extremely biased populations with a single represented group, and approaches $0$ for increasingly diverse populations, with a lower bound of $1/\text{R}(X)$. Two variants are commonly employed, the \textit{Simpson’s Index of Diversity} defined by $1-\text{D}(X)$, and the \textit{Simpson's Reciprocal} defined as $1/{\text{D}(X)}$. Of these three, Simpson's Reciprocal index is of particular interest in this context, as it shares the same range of ENS, from $1$ in populations representing biased in favor of a single group to an upper limit of $\text{R}(X)$ for more diverse populations. This metric is more influenced by the \textit{evenness} of the population compared to ENS, especially when there are more groups present.

\subsection{Metrics for stereotypical bias}\label{ssec:stereo_metrics}

The following metrics can be used to measure stereotypical bias from a \textbf{global} perspective.

\vspace{0.3em}\noindent\textbf{Cramer's V} ($\phi_C$)~\cite{Cramer1991}, \textbf{Tschuprow's T} (T)~\cite{Tschuprow1939} and \textbf{Pearson's Contingency Coefficient} (C)~\cite{Sakoda1977}. These three metrics are directly based on the Pearson's chi-squared statistic of association (${\chi^2}(X)$), employed in the popular \textit{Pearson's chi-squared test}. The ${\chi^2}(X)$ statistic is defined as:
\begin{equation}
    \chi^2(X)=\sum_{g \in G}\sum_{y \in Y}\frac{(n_{g\land y}-\frac{n_g n_y}{n})^2}{\frac{n_g n_y}{n}}\;.
\end{equation}

As the ${\chi^2}(X)$ calculates the difference between the real number of samples of a subgroup $n_{g\land y}$ and the expected number of samples of the subgroup $\frac{n_g n_y}{n}$, it detects the under or overrepresentation of specific subgroups independently of the potential representational bias in the distribution of target classes $y \in Y$ or demographic groups $g \in G$. Unfortunately, the result is both dependent on the total number of samples $n$ and does not have clear units or intuitive bounds, making it difficult to interpret as a bias metric. Due to this, several corrections with defined bounds are available. In particular, $\phi_C(X)$, $T(X)$, and $C(X)$ are defined as:
\begin{equation}
  \phi_C(X) = \sqrt{ \frac{\chi^2(X)/n}{\min(|G|-1,|Y|-1)}}\;,
\end{equation}
\begin{equation}
    \text{T}(X) = \sqrt{ \frac{\chi^2(X)/n}{\sqrt{(|G| - 1) \cdot (|Y| - 1)}}}\;, \text{and}
\end{equation}
\begin{equation}
    \text{C}(X) = \sqrt{ \frac{\chi^2(X)/n}{{1 - \chi^2(X)/n}}}\;.
\end{equation}

The three metrics share the same bounds, from $0$, which represents no bias or association between the demographic component and the target class, up to $1$, a maximum bias or association. This difference makes them generally more meaningful and interpretable than the original statistic. Both $\text{T}(X)$ and $\text{C}(X)$ can only achieve their theoretical maximum of $1$ when both nominal variables have the same number of possible values, $|G| = |Y|$. This restriction does not apply to $\phi_C(X)$, whose notion of correlation can be maximized even when $|G| \neq |Y|$. This difference makes $\phi_C(X)$ more widely used in practice.

Additionally, thresholds of significance have been provided for $\phi_C(X)$~\cite{Cohen1988}, and can be also used when measuring bias. These thresholds depend on the degrees of freedom of the metric, calculated as $\text{DoF}(X) = \min(|G|-1,|Y|-1)$ for our application. In particular, for $\text{DoF}(X)=1$, $\phi_C < 0.1$ is considered a small or weak association or bias, $\phi_C < 0.3$ a medium bias, and $\phi_C < 0.5$ a large or strong bias. For $\text{DoF}(X)>1$, the thresholds are corrected to $0.1 / \sqrt{\text{DoF}(X)}$, $0.3 / \sqrt{\text{DoF}(X)}$, and $0.5 / \sqrt{\text{DoF}(X)}$, respectively.

\vspace{0.3em}\noindent\textbf{Theil's U} (U)~\cite{Theil1970} or \textit{Uncertainty Coefficient} is a measure of association based on Shannon entropy, that can therefore be employed to measure stereotypical bias. It is defined as:
\begin{equation}
    \text{U}(X, P_1 \rightarrow P_2) = \frac{\text{H}(X, P_1) - \text{H}(X, P_1|P_2)}{\text{H}(X, P_2)}\;,
\end{equation}
where $P_1$ and $P_2$ stand for $G$ and $Y$ in any order. In this paper, we will treat $P_1=G , P_2 = Y$ as the \textit{default} order (denoted by $\text{U}(X)$) and $P_1=Y , P_2 = G$ as the \textit{reverse} order (denoted by $\text{U}^\text{R}(X)$). Additionally, $\text{H}(X, P)$ with $P \in \{P_1, P_2\}$ is defined as:
\begin{equation}
    \text{H}(X, P) = -\sum_{i\in P} p_i \ln(p_i)\;,
\end{equation}
and $\text{H}(X, P_1|P_2)$ is defined as:
\begin{equation}
    \text{H}(X, P_1|P_2) = -\sum_{i \in P_1} \sum_{j\in P_2} 
    p_{i\land j} \ln\left(\frac{p_{i \land j}}{p_j}\right)\;.
\end{equation}

This metric has a lower bound of $0$, no bias or association between the demographic component and the target class, and an upper bound of $1$, a maximum bias or association. 

The definitions of stereotypical bias presented up to this point are all direction-agnostic, meaning that they produce the same result for any pair of variables, regardless of which one is provided first. A key characteristic of the Theil's U metric is that it is instead asymmetric, measuring the proportion of uncertainty reduced in one of the variables (target) when the other one (source) is known. Thus, the application of this metric could potentially establish a differentiation between forward and backward stereotypical bias.

\vspace{0.3em}\noindent\textbf{Normalized Mutual Information} (NMI)~\cite{Dominguez-Catena2022,Bouma2009}. A different approach to measuring the association between two variables is the use of Mutual Information based variables. In particular, a previous work~\cite{Dominguez-Catena2022} used a normalized variant to measure stereotypical bias in a dataset.

\begin{equation}
    \text{NMI}(X) = -\frac{ \displaystyle\sum_{g \in G} \displaystyle\sum_{y \in Y} p_{g \land y} \ln\frac{p_{g \land y}}{p_g p_y}} {\displaystyle\displaystyle\sum_{g \in G} \displaystyle\sum_{y \in Y} p_{g \land y} \ln{p_{g \land y}}}\;.
\end{equation}

The value of $\text{NMI}(X)$ is in the range $[0, 1]$, with $0$ being no bias and $1$ being total bias. 

Finally, the stereotypical bias can also be measured using the following \textbf{local} metrics.

\vspace{0.3em}\noindent\textbf{Normalized Pointwise Mutual Information} (NPMI)~\cite{Dominguez-Catena2022,Bouma2009}. The $\text{NMI}(X)$ metric has a local variant, NPMI. This metric has a different application than the previous metrics, as it is not intended for the analysis of the stereotypical bias in the dataset as a whole. Instead, NPMI is a local stereotypical bias metric capable of highlighting the particular combination of demographic groups and the target class in which bias is found. Mathematically, it is defined as:

\begin{equation}
    \text{NPMI}(X, g, y) = -\frac{\ln\frac{p_{g\land y}}{p_g p_y}} {\ln p_{g \land y}}\;,
\end{equation}
where $g \in G$ is the demographic group of interest and $y \in Y$ is the target class. The values of $\text{NPMI}$ are in the range $[-1, 1]$, with $1$ being the maximum overrepresentation of the combination of group and class, $0$ being no correlation and $-1$ being the maximum underrepresentation of the combination.

\vspace{0.3em}\noindent\textbf{Ducher's Z} (Z)~\cite{Ducher1994}. The $\text{Z}$ measure of local association, originally developed in the field of biology, can also be employed to measure local stereotypical bias. It is defined as:

\begin{equation}
    \text{Z}(X, g, y) = 
      \begin{cases}
        \frac{p_{g\land y}-p_g p_y}{\min[p_g, p_y]-p_g p_y} & \text{if } p_{g\land y} - p_g p_y > 0\\ 
        \frac{p_{g\land y}-p_g p_y}{p_g p_y - \max[0, p_g+ p_y-1]} & \text{if } p_{g\land y} - p_g p_y < 0\\ 
        0 & \text{otherwise,}\\ 
      \end{cases}
\end{equation}
where $g \in G$ is the demographic group of interest and $y \in Y$ is the target class. The values of $\text{Z}(X)$ are also in the range $[-1, 1]$, with $1$ being the maximum overrepresentation of the combination of group and class, $0$ being no correlation and $-1$ being the maximum underrepresentation of the combination.

\section{Case study: FER datasets and demographic information}\label{sec:experiments}

In this section, we present the experimental framework used to observe the real-world behavior of bias metrics in the FER case study. First, Section~\ref{ssec:datasets} presents the selection of datasets used in this work. Then Section~\ref{ssec:preprocessing} details the steps taken to preprocess and homogenize the different datasets. Finally, Section~\ref{ssec:fairface} explains the demographic profiling of the samples in the datasets, as a necessary step to enable the application of the bias metrics.

\subsection{Datasets} \label{ssec:datasets}

For this work, we initially considered a total of $55$ datasets used for FER tasks, collected from a combination of dataset lists provided in previous reviews~\cite{Guerdelli2022,Li2020,Guerdelli2022}, datasets cited in various works, and datasets discovered through Internet searches. This list was further pruned to a final list of $20$ datasets, presented in Table~\ref{table:datasets}, according to the following criteria:

\begin{enumerate}
    \item 2D image-based datasets, or video-based datasets with per-frame labeling. This is the most extended approach to FER.
    \item Datasets based on real images. Although some artificial face datasets are available, the demographic relabeling process can be unreliable in these contexts.
    \item Datasets that include labels for the six basic emotions (anger, disgust, fear, happy, sadness, and surprise), and optionally neutral. This codification is the most popular in FER datasets, and the adoption of a unified label set makes the stereotypical biases comparable across datasets.
    \item Availability of the datasets at the time of request.
\end{enumerate}

\begin{table*}[ht]
  \begin{center}
    \caption{Summary of the FER Datasets and their characteristics.}
    \label{table:datasets}
    \resizebox{\textwidth}{!}{
    \pgfkeys{/pgf/number format/fixed}
    \begin{filecontents*}{data/datasets.csv}
Abbreviation,"Full Name",Year,Collection,Images,Videos,Subjects,Labelling$^a$
"ADFES \cite{vanderSchalk2011}","Amsterdam Dynamic Facial Expressions Set",2016,LAB,---,648,22,9
"AffectNet \cite{Mollahosseini2019}",AffectNet,2017,ITW,"291,652",---,---,"7 + N"
"CAER-S \cite{Lee2019}","CAER Static",2019,ITW-M,"70,000",---,---,"6 + N"
"CK \cite{Kanade2000}","Cohn-Kanade Dataset",2000,LAB,"8,795",486,97,"7 + N"
"CK+ \cite{Lucey2010}","Extended Cohn-Kanade Dataset",2010,LAB,"10,727",593,123,"7 + N + FACS"
"ExpW \cite{Zhang2017}","Expression in-the-Wild",2017,ITW,"91,793",---,---,"6 + N"
"FER2013 \cite{Goodfellow2013}","Facial Expression Recognition 2013",2013,ITW,"32,298",---,---,"6 + N"
"FERPlus \cite{Barsoum2016}","FER2013 Plus",2016,ITW,"32,298",---,---,"6 + N"
"GEMEP \cite{Banziger2012}","Geneva Multimodal Emotion Portrayals",2010,LAB,"2,817","1,260",10,17
"iSAFE \cite{Singh2020}","Indian Semi-Acted Facial Expression",2020,LAB,---,395,44,"6 + N + U"
"JAFFE \cite{Lyons1998, Lyons2021}","Japanese Female Facial Expression",1998,LAB,213,---,10,"6 + N"
"KDEF \cite{Lundquist1998}","Karolinska Directed Emotional Faces",1998,LAB,"4,900",---,70,"6 + N"
"LIRIS-CSE \cite{Khan2019}","LIRIS Children Spontaneous Facial Expression Video Database",2019,LAB,"26,000",208,12,"6 + U"
"MMAFEDB \cite{_MMAFACIALEXPRESSION}","Mahmoudi MA Facial Expression Database",2020,ITW,"128,000",---,---,"6 + N"
"MUG \cite{Aifanti2010}","Multimedia Understanding Group Facial Expression Database",2010,LAB,"70,654",---,52,"6 + N"
"NHFIER \cite{_NaturalHumanFace}","Natural Human Face Images for Emotion Recognition",2020,ITW,"5,558",---,---,"7 + N"
"Oulu-CASIA \cite{Zhao2011}",Oulu-CASIA,2008,LAB,"66,000",480,80,6
"RAF-DB \cite{Li2017, Li2019}","Real-world Affective Faces Database",2017,ITW,"29,672",---,---,"6 + N"
"SFEW \cite{Dhall2011}","Static Facial Expressions In The Wild",2011,ITW-M,"1,766",---,330,"6 + N"
"WSEFEP \cite{Olszanowski2015}","Warsaw Set of Emotional Facial Expression Pictures",2014,LAB,210,---,30,"6 + N + FACS"
\end{filecontents*}
\pgfplotstabletypeset[
        col sep=comma,
        text indicator=",
        columns={Abbreviation, 
            [index]1, 
            Year, 
            Collection, 
            Images, 
            Videos, 
            Subjects, 
            [index]7},
        sort=true,
        sort key=Year,
        string type,
        display columns/0/.style={column type = {l}},
        display columns/1/.style={column type = {l}},
        display columns/2/.style={column type = {r}},
        display columns/3/.style={column type = {l}},
        display columns/4/.style={column type = {r}},
        display columns/5/.style={column type = {r}},
        display columns/6/.style={column type = {r}},
        display columns/7/.style={column type = {l}},
        every head row/.style={
            before row=\toprule,after row=\midrule},
        every last row/.style={
            after row=\bottomrule}
    ]{data/datasets.csv}}
  \end{center}
 \footnotesize{$^a$ 6: angry, disgust, fear, sad, surprise, and happy. 7: 6 + contempt. N: Neutral. U: Uncertain. FACS: Facial Action Coding System.}
\end{table*}

These datasets can be categorized into three groups, depending on the source of the images:

\begin{itemize}
    \item \textbf{Laboratory-gathered} (\textit{Lab}), which usually includes a limited selection of subjects whose images or sequences of images are taken under controlled conditions. The images in these datasets are intended for FER from inception, so the images are usually high-quality and taken in consistent environments.
    \item In The Wild from \textbf{Internet queries} (\textit{ITW-I}). These datasets are created from images not intended for FER learning, so their quality is varied. These datasets usually have a larger number of images, as their sourcing is relatively cheap.
    \item In The Wild from \textbf{Motion Pictures} (\textit{ITW-M}). These datasets try to improve the inconsistent quality of \textit{ITW-I} datasets by sampling their images from motion pictures (including video from films, TV shows, and other multimedia), while retaining the advantages of a relatively high number of samples.
\end{itemize}

\subsection{Data preprocessing} \label{ssec:preprocessing}

To enable the comparison of the studied datasets, we preprocessed them to make the data as homogeneous as possible and to ensure accurate demographic labeling. For every dataset, we performed the following steps:

\begin{enumerate}
\item \textbf{Frame extraction.} For datasets based on video, namely, ADFES, CK, CK+, GEMEP, iSAFE, LIRIS-CSE, and Oulu-CASIA, we either extracted the frames or used the per frame version when available. 

\item \textbf{Face extraction.} Although most of the datasets provide extracted face images, the resolution and amount of margin around the face often vary considerably. To facilitate demographic prediction (see Section~\ref{ssec:fairface}), we used the same face extraction methodology of FairFace~\cite{Karkkainen2021}, namely a Max-Margin (MMOD) CNN face extractor~\cite{King2015} implemented in DLIB\footnote{\url{http://dlib.net/}}. Face extraction was performed with a target face size of $224 \times 224$ (resized if necessary) with a margin around the face of $0.25$ ($56$ pixels). When needed, a zero padding (black border) was used when the resized image includes portions outside the original image. On the EXPW dataset, where images have several faces, we applied the same process to each individual face extracted from the face bounding boxes provided.

\item \textbf{Emotion classification relabeling.} For each dataset, we consider the images corresponding to the six basic emotions~\cite{Ekman1971}: angry, disgust, fear, sad, surprise, and happy, plus a seventh category for neutrality. For some datasets, the original emotion names differ (such as \textit{angry} and \textit{fury}). In these cases, we only rename the emotion if it is a clear synonym of the intended one. Additionally, it must be noted that not all included datasets provide examples of all emotions, with some (Oulu-CASIA and GEMEP) missing examples of neutrality.

\end{enumerate}

\subsection{Fairface} \label{ssec:fairface}

For the analysis of demographic bias in datasets, it is indispensable to have demographic information of the depicted subjects. In the context of FER, the large majority of datasets do not provide or gather this information, or when they do, it is often partial infomation (such as in ADFES, which only refers to race and gender) or global statistics for the whole dataset and not to each sample (such as CK+). In most ITW datasets, in particular those based on Internet queries, this information is unavailable even to the original developers, as the subjects are mostly anonymous (such as in FER2013).

To overcome this limitation, we propose, following our previous work~\cite{Dominguez-Catena2022,Dominguez-Catena2023}, the study of biases with respect to a proxy demographic prediction instead of the original unrecoverable information. This can be achieved through a secondary demographic model, such as the FairFace~\cite{Karkkainen2021} face model, based on the homonymous dataset. The FairFace model is trained with the FairFace dataset, made up of $108,501$ images of faces from Flickr (an image hosting service) hand-labeled by external annotators according to apparent race, gender, and age. The dataset is designed to be balanced across seven racial groups, namely White, Black, Indian, East Asian, Southeast Asian, Middle Eastern, and Latino. The dataset is also labeled with both binary gender (Male or Female) and age group (9 age groups), although not fully balanced across these characteristics. The trained model is publicly available\footnote{\url{https://github.com/joojs/fairface}} and was studied against other demographic datasets, namely UTKFace, LFWA+ and CelebA, improving previous results and showing accurate classification in gender and race categories, and more modest results in the age category.

It is important to note that even for FairFace, the demographic data comes from external annotators and not self-reported demographic characteristics. Furthermore, even self-reported race and gender identities are extremely subjective and for many individuals complicated to specify~\cite{TheGenIUSSGroup2014}. This labeling is also limited to a small set of fixed categories, leaving out many possible race and age descriptors~\cite{Keyes2018}. The lack of descriptors means that any classification based on these categories is fundamentally incomplete, leaving out individuals from potentially impacted minorities and misrepresenting the true identity of others.

Despite these limitations, in problems like FER, the system can only discriminate based on apparent age, gender, and race, as no true demographic information is available. Therefore, the same apparent categories can be used in our bias analysis. In such cases, FairFace provides a useful prediction of these categories, enabling a study that would otherwise be impossible.

\section{Results}\label{sec:results}

The main objective of this section is to evaluate the behavior of dataset bias metrics when evaluating real-world datasets, in our case targeted to the FER task. To this end, we aim to answer the following research questions:

\begin{enumerate}
  \item Do the different representational dataset bias metrics experimentally agree with each other? What metrics constitute the minimal and sufficient set that can characterize representational bias?
  \item Do the different global stereotypical dataset bias metrics experimentally agree with each other? What metrics constitute the minimal and sufficient set that can characterize global stereotypical bias?
  \item Do the local stereotypical dataset bias metrics experimentally agree? What is the most interpretable local stereotypical bias metric?
  \item How much representational and stereotypical bias can be found using the aforementioned metrics in FER datasets? Which FER datasets are the most and less biased?
\end{enumerate}

In this section, we present the main results of the case study, over the selected datasets and the three demographic components, namely age, race, and gender. First, Sections~\ref{ssec:agreement-rep},~\ref{ssec:agreement-stereo} and~\ref{ssec:agreement-local} focus on the agreement results of the metrics in the categories of representational, global stereotypical, and local stereotypical bias, respectively. Next, Section~\ref{ssec:bias_in_fer} presents the bias analysis of the datasets with respect to representational and stereotypical bias.

\subsection{Agreement between representational bias metrics}\label{ssec:agreement-rep}

\begin{figure*}[hbt]
    \centering
    \includegraphics[width=\textwidth]{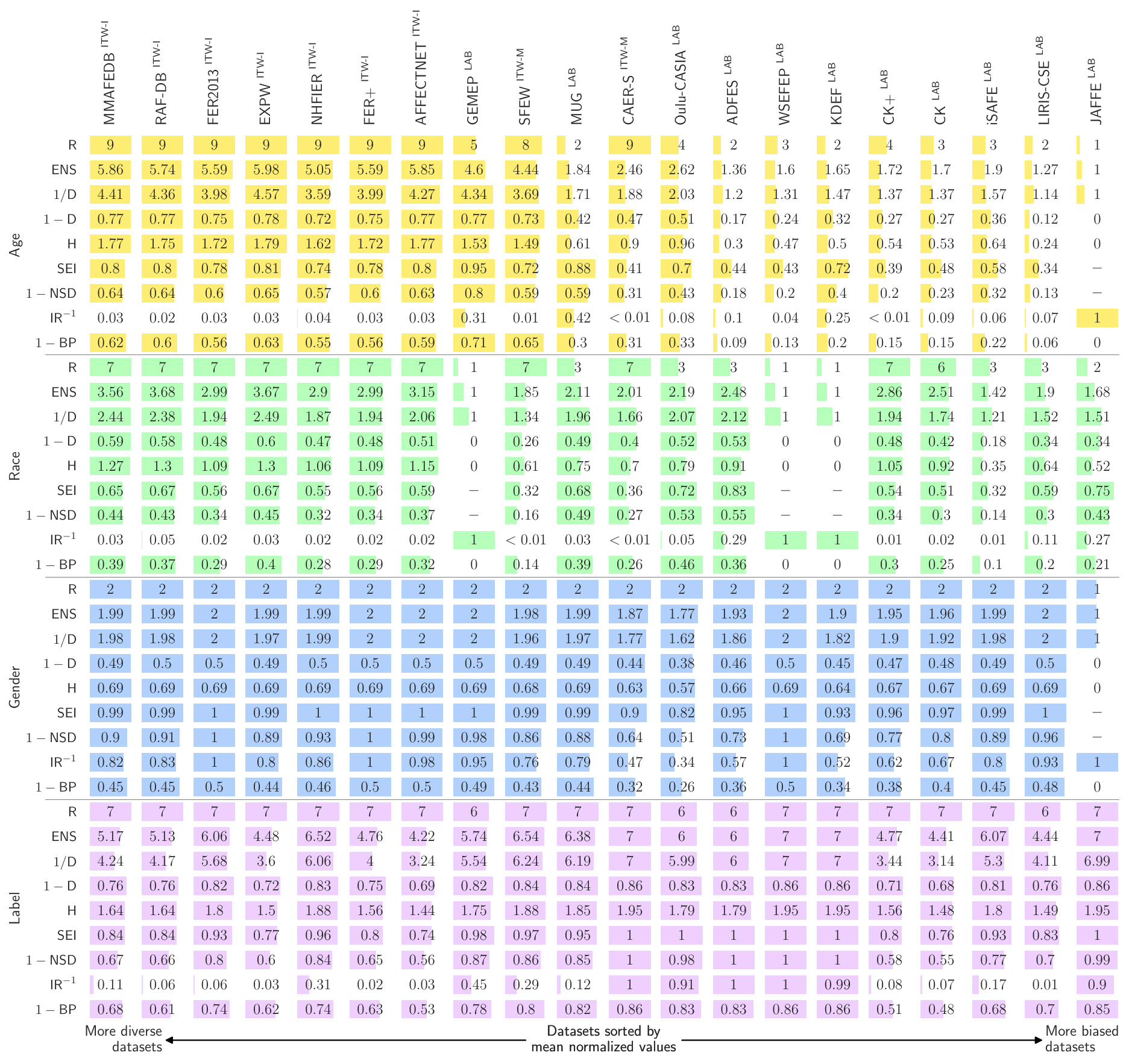}
    \caption{Representational bias metrics for the the three demographic components and the target label. The metrics are calculated as diversity metrics, with higher values corresponding to lower representational bias. The graphical representations of the values are normalized to the maximum value of the row. The datasets are sorted by the average of the normalized metrics.}
    \label{figure:rep_bias_detail}
\end{figure*}

\noindent\textbf{Overview.} Figure~\ref{figure:rep_bias_detail} shows the values obtained for the different representational bias metrics presented in Section~\ref{ssec:rep_metrics} when applied to each of the datasets considered, in each of the three potential demographic axes. Additionally, we apply the metrics to the distribution of the target label, as an example of their use on non-demographic components. As most metrics in this category are diversity metrics, bias metrics, such as D, NSD, and BP, are computed in their complementary form based on their upper limit $1$, that is, $1-\text{D}$, $1-\text{NSD}$ and $1-\text{BP}$, respectively, to allow for a more direct comparison. IR is computed in its inverse form, $\text{IR}^{-1}$, as it has a closer range to the other metrics. For a better visualization, the plotted values are normalized by the maximum value of each line (metric in each component). The datasets are sorted by the average of these normalized values in decreasing order, from higher average values, which indicate less representational bias, to lower values which indicate more representational bias.

Globally, we can observe that the ranking of the datasets according to the different metrics is mostly consistent, suggesting a high agreement between the metrics even if the scales vary. A marked exception occurs in the combinations of dataset and component where a single group is represented ($\text{R}(X)=1$), such as JAFFE in the age and gender components, and GEMEP, WSEFEP, and KDEF, in the race component, where $\text{IR}^{-1}$ reports high values, corresponding to a situation of high diversity and low bias, contrary to the intuitive notion that these datasets are strongly biased in these components. These results are caused by a strict implementation of the IR metric, since in these datasets and components the most and least represented groups are the same. Additionally, it can be observed that $\text{IR}^{-1}$ has a different behavior in the gender component, where the values are similar to those of the other metrics, compared to the other components. This can be explained by the number of groups in each component, with only two groups in the gender component while the rest of the components have more, as well as by the balance in the components, with the gender component showing less representational bias overall. In these cases, the limitations of the $\text{IR}^{-1}$ metric have less impact in this context, when applied to components with few groups represented in roughly equal proportions. 

Similar to the behavior of $\text{IR}^{-1}$, some of the metrics, such as SEI and NSD, are directly ill-defined in the the trivial cases where a single demographic group is present in the dataset. Metrics related to R, such as ENS, $1/\text{D}$ or R itself, are instead robust to this cases, simply confirming the intuition that only one group is represented, and giving a result always lower than for datasets with more groups represented, no matter their evenness.

\vspace{0.3em}\noindent\textbf{Interpretability.} To analyze the interpretability of these metrics one of the key factors is the range of the metric, as the bounds contextualize the values of a metric. Some of the metrics, such as R, ENS, and $1/\text{D}$, are bounded in a $[1,R]$ range, immediately interpretable as a number of represented groups. From these, ENS and $1/\text{D}$ complement the pure richness information with the evenness of the population, lowering the value when some of the groups are underrepresented. In the case of metrics focused on a single characteristic of representational bias, such as evenness (SEI, NSD) and dominance (BP, $\text{IR}^{-1}$), the most common range is $(0,1)$ (including or excluding the bounds). This unitary range is easy to interpret for these characteristics, and allows for a quick conversion between diversity and bias when one of the meanings is preferred, as we have done with $D$, $NSD$ and $BP$. Among the metrics associated with this unitary range, we can observe that $\text{IR}^{-1}$ tends to exaggerate the biases compared to the other metrics in components with more than two groups (all but gender). This results in $\text{IR}^{-1}$ values close to 0 in these components for most datasets. Finally, the H value lies in the range $[0,\ln\left(\text{R}(X)\right)]$, with no natural interpretation. As the information conveyed by H is the same as the one in the ENS, while being less interpretable, ENS is generally preferable.

\begin{figure}[hbt]
    \centering
    \includegraphics[width=0.95\columnwidth]{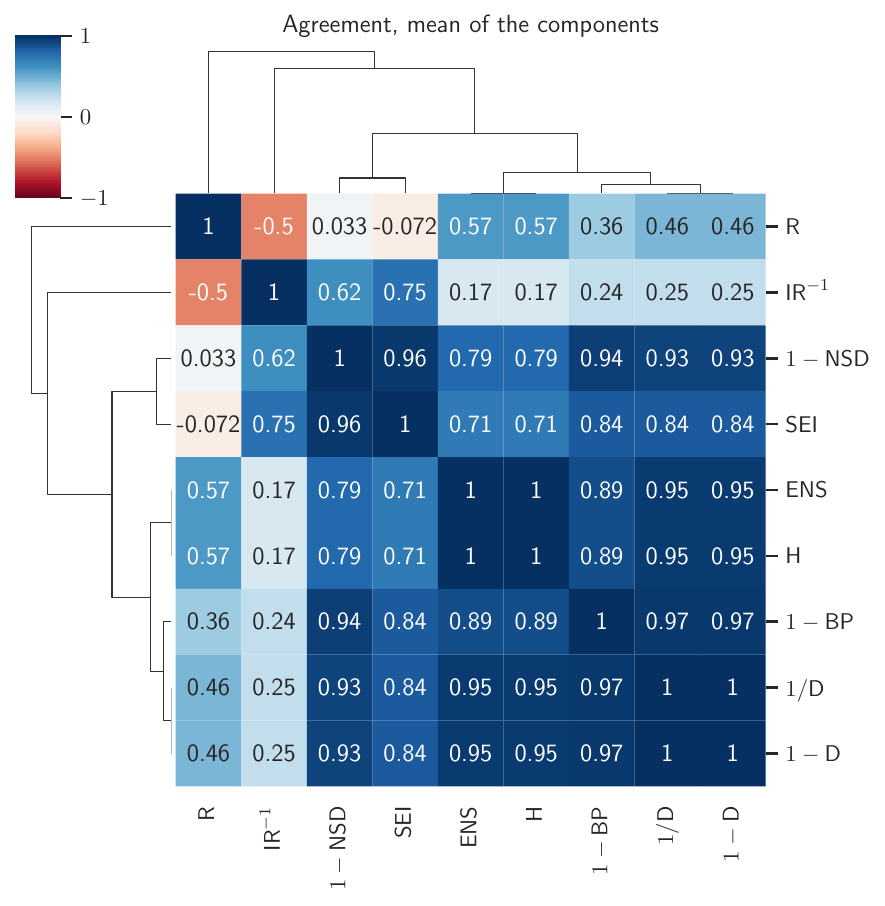}
    \caption{Spearman's $\rho$ agreement between the representational bias metrics, measured independently for each component and then averaged for each pair of metrics. Higher $\rho$ values indicate high coherence between the rankings generated by the metrics.}
    \label{figure:agre_representational}
\end{figure}

\vspace{0.3em}\noindent\textbf{Statistical agreement assessment.} To evaluate the agreement between the metrics, we employ the Spearman's $\rho$, a measure of the strength and direction of the monotonic relationship of two variables interpreted as ordinal, that is, where only the ranking produced is considered. To compute the pairwise agreement between metrics, we employ Spearman's $\rho$ to compare each pair of metrics on each component (using the results in Figure~\ref{figure:rep_bias_detail}) and then average across the four components. This procedure highlights the similarity across the metrics. Pairs of metrics that order the bias in the same way (indicating redundancy among the metrics) produce values of $\rho$ close to $1$ or $-1$, depending on the direction of the relationship, and less related metrics produce values close to $0$. Figure~\ref{figure:agre_representational} shows these agreement results between the different metrics of representational bias.

Except for R and $\text{IR}^{-1}$, the agreement value is greater than $0.71$ for most metrics, with an average of $0.88$. In the case of R, we can observe very low correlations with the rest of the metrics. The low robustness of this metric, where adding a single example of a missing demographic group to a whole dataset will produce a change in its value, accounts for these low correlations. In the same way, $\text{IR}^{-1}$ is especially unreliable on demographic axes with many potential groups, such as race and gender, becoming overly sensitive to the representation of the least represented group.

Outside of these two exceptions, the rest of the metrics mostly conform to the taxonomy presented in Section~\ref{ssec:taxonomy}. SEI and NSD are strongly correlated ($\rho = 0.96$), as both measure evenness, and both of them are, as expected, independent of R, with $|\rho| < 0.1$ in the two cases. The other cluster of metrics with high correlation is mainly made up of combined metrics, namely, ENS, H, $1/\text{D}$ and $1-\text{D}$. Unexpectedly, the other dominance metric, BP, is also included in this cluster, with high correlation with the others (between $0.85$ and $0.96$). This metric, different from $\text{IR}^{-1}$ in that it does not directly consider the least represented group, is more reliable and appears to experimentally capture information similar to the rest of the metrics. Considering the simplicity and high interpretability of BP, this makes it an interesting metric for general representational bias characterization.

It is also noteworthy the high correlation between the combined and evenness metrics clusters, showing how in this particular case of study both components of representational bias are highly related.

\vspace{0.3em}\noindent\textbf{Recommended metrics.} In general, characterizing the bias of a dataset for a given component appears to be appropriately summarized by one of the combined metrics bounded by R, namely, ENS and $1/\text{D}$, as these combined metrics are highly interpretable variants of the pure richness, and one of the evenness metrics to highlight this independent component, such as SEI or NSD. For more succinct analysis where evenness and richness are not expected to substantially disagree, Berger-Parker appears to maintain a high correlation with other metrics while having an intuitive meaning and simple implementation. 

\subsection{Agreement between stereotypical bias metrics}\label{ssec:agreement-stereo}

\begin{figure*}[hbt]
    \centering
    \includegraphics[width=\textwidth]{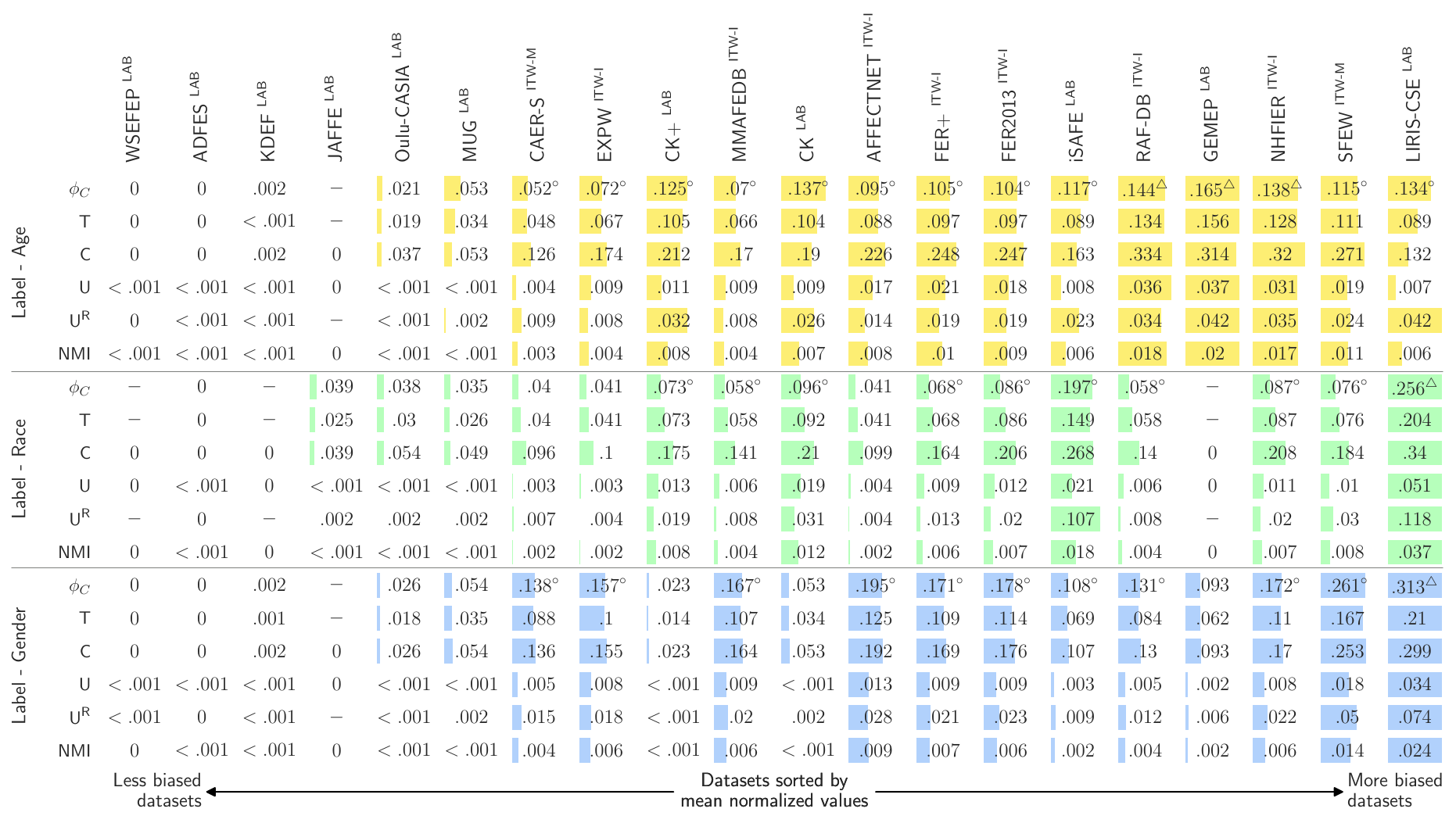}
    \caption{Stereotypical bias metrics for the the three demographic components against the target label. Higher values correspond to higher amounts of stereotypical bias. The graphical representation at each row, corresponding to a single metric and demographic component, is normalized to the maximum value of the row. In the $\phi_C$ row a ${}^\circ$ mark indicates a statistically weak association and a ${}^\triangle$ mark a statistically medium association. {The datasets are sorted by the average of the normalized metrics, from lower values (less stereotypical bias) in the left to higher values (more stereotypical bias) in the right.}}
    \label{figure:stereo_bias_detail}
\end{figure*}

\noindent\textbf{Overview.} Figure~\ref{figure:stereo_bias_detail} shows the stereotypical bias metric results for the three main demographic components against the target label. These metrics target bias directly, thus, higher values, shown in the right, relate to more stereotypically biased datasets. The graphical representation of the values is normalized by the maximum value of each line (metric in each component). Following the thresholds for $\phi_C$ presented in Section~\ref{ssec:stereo_metrics}, we indicate the bias strength in the row $\phi_C$. ${}^\circ$ marks a weak bias, and ${}^\triangle$ marks a medium bias. No strong bias are found according to these thresholds.

The ranking produced by these stereotypical bias metrics is highly coherent, especially in the gender component, where a binary classification is used. According to the magnitude of the values reported, two different groups are observed. The three metrics based on the $\chi^2$ metric, namely, $\phi_C$, T, and C, all report values in a similar range for all datasets. The NMI and the U metric, applied both in the forward and reverse direction, noted here as $\text{U}^\text{R}$, tend to report coherent but lower values than the other three metrics. U and $\text{U}^\text{R}$ result in similar values and rankings, with a few exceptions, namely, the CK+, CK, iSAFE, and LIRIS-CSE datasets in the age component, and the iSAFE dataset in the race component.

\vspace{0.3em}\noindent\textbf{Interpretability}. All of the stereotypical bias metrics lie in the range $[0,1]$ and share the same interpretation of these bounds, namely, $0$ for no bias and $1$ for maximal bias, which allow for a simple interpretation. The distribution of the values, as mentioned earlier, is different, with U, $\text{U}^\text{R}$ and NMI resulting in very low values for all of the datasets, which is indicative of a low sensitivity. The three other metrics, $\phi_C$, T and C, all show higher values, related to a higher sensitivity. Additionally, for $\phi_C$ both the availability of predefined thresholds and the fact that it can be maximized even when the number of demographic groups and target labels is not equal (see Section~\ref{ssec:stereo_metrics}) improve its interpretability.

\begin{figure}[hbt]
    \centering
    \includegraphics[width=0.72\columnwidth]{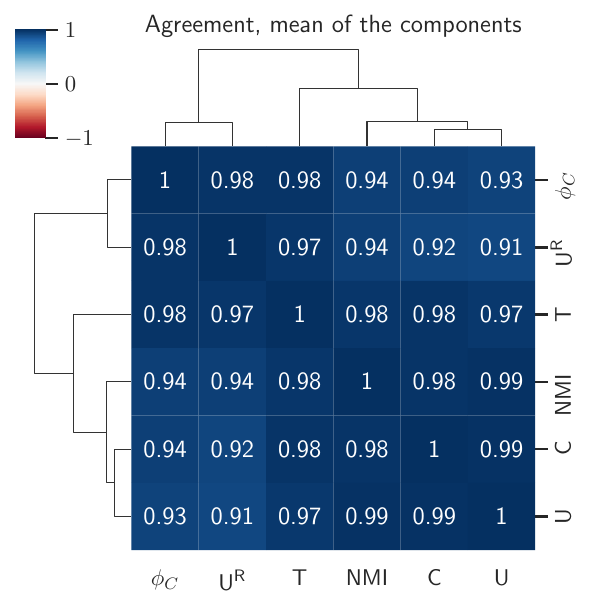}
    \caption{Spearman's $\rho$ agreement between the stereotypical bias metrics, measured independently for each component and then averaged for each pair of metrics. Higher $\rho$ values indicate high coherence between the rankings generated by the metrics.}
    \label{figure:agre_stereotypical}
\end{figure}

\vspace{0.3em}\noindent\textbf{Statistical agreement assessment}. Figure~\ref{figure:agre_stereotypical} uses Spearman's $\rho$ to compare the agreement between the different stereotypical bias metrics, measured between the three main demographic components and the output label. As intuitively observed in Figure~\ref{figure:stereo_bias_detail}, the agreement values are high in all cases, with a minimum of $0.91$, between U and $\text{U}^\text{R}$. It can be concluded that for this case study all metrics provide similar information about the amount of information shared between the three demographic components and the output classes.

\vspace{0.3em}\noindent\textbf{Recommended metrics.} We observe no notable differences or clusters in the stereotypical bias metrics. As the ranking of biases is experimentally similar, any of the metrics is sufficient for comparing the bias between datasets. In spite of this, we suggest using $\phi_C$ for stereotypical bias characterization, with a more natural range of values and predefined thresholds for determining the intensity of the bias, providing a more interpretable result. 

\subsection{Agreement between local stereotypical bias metrics}\label{ssec:agreement-local}

\begin{figure*}[hbt]
    \centering
    \begin{minipage}{0.29\textwidth}
        \includegraphics[width=\columnwidth]{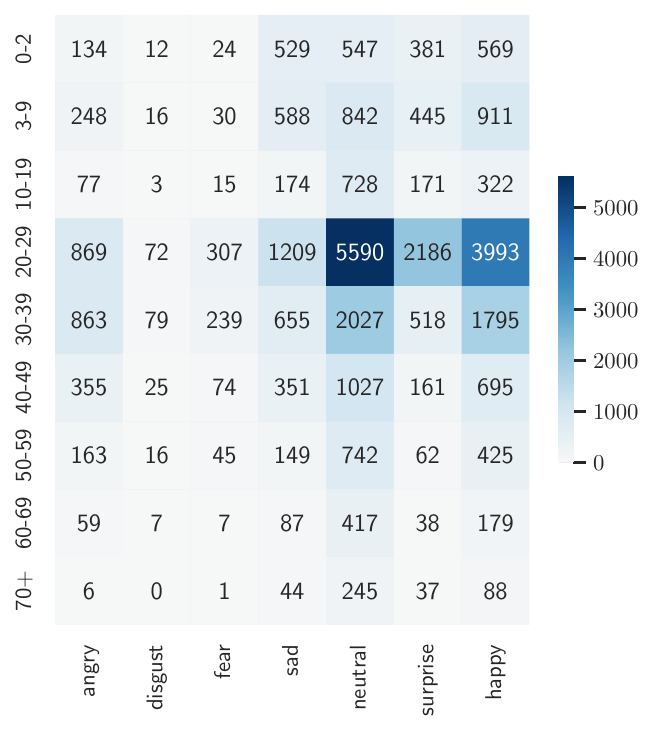}
        \includegraphics[width=\columnwidth]{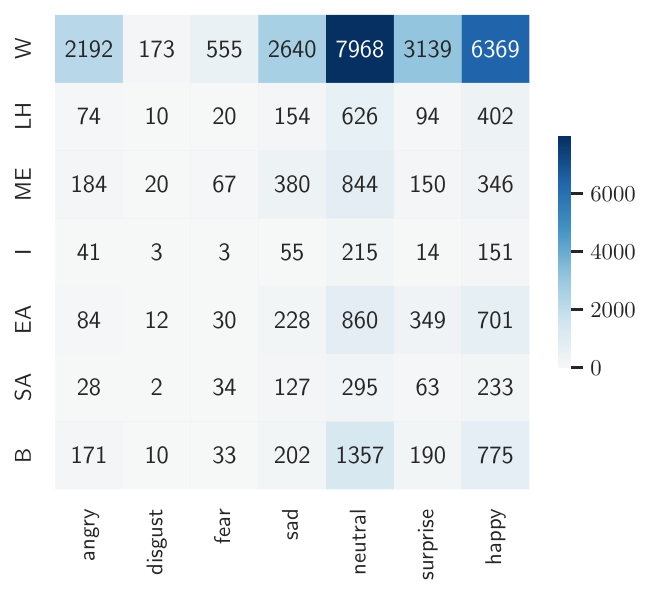}
        \includegraphics[width=.99\columnwidth]{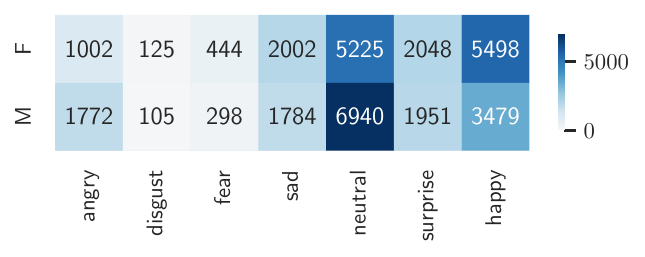}
    \end{minipage}%
    \begin{minipage}{0.29\textwidth}
        \includegraphics[width=\columnwidth]{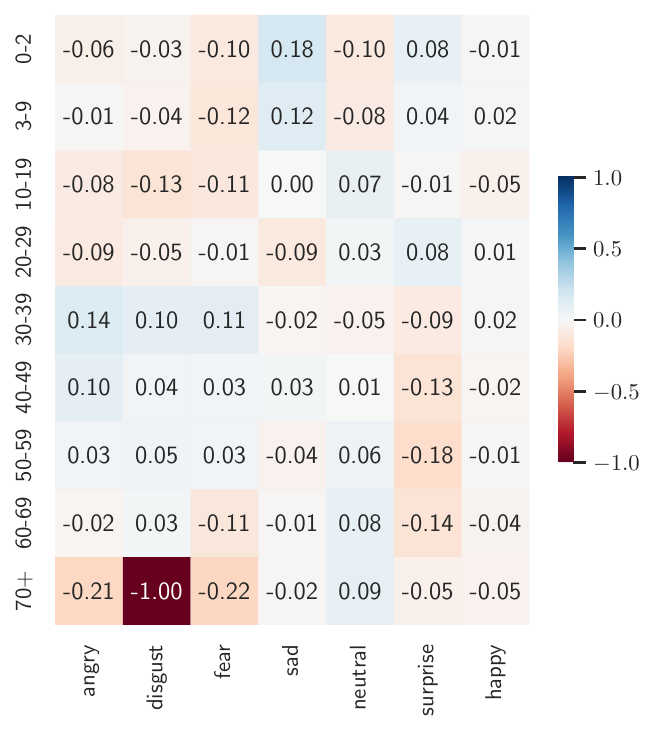}
        \includegraphics[width=\columnwidth]{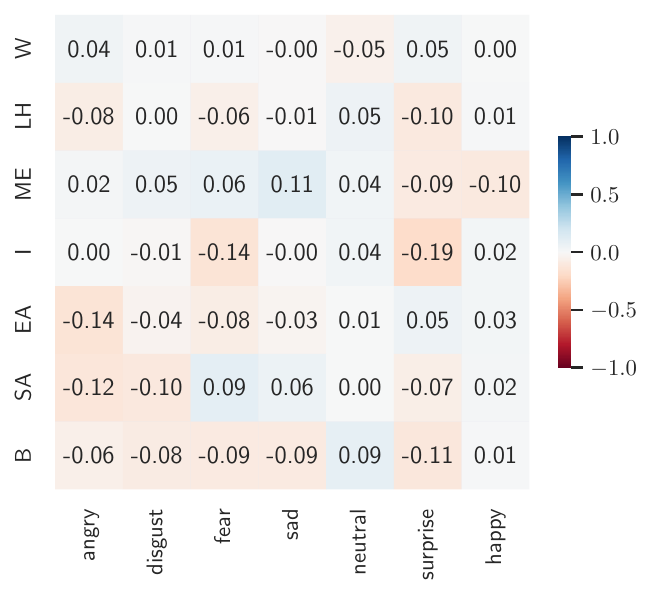}
        \includegraphics[width=.96\columnwidth]{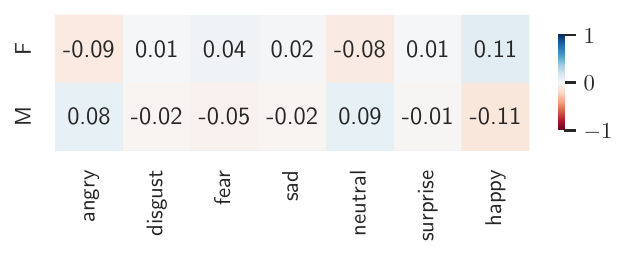}
    \end{minipage}%
    \begin{minipage}{0.29\textwidth}
        \includegraphics[width=\columnwidth]{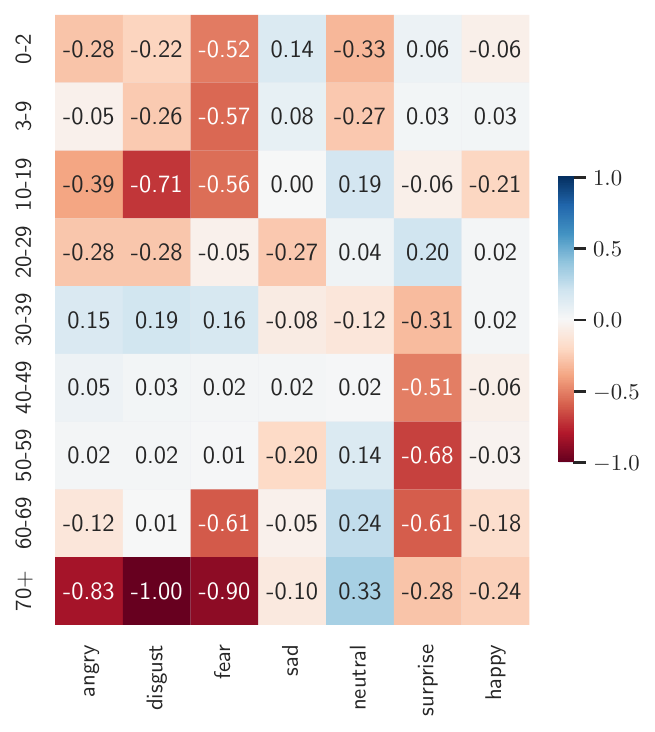}
        \includegraphics[width=\columnwidth]{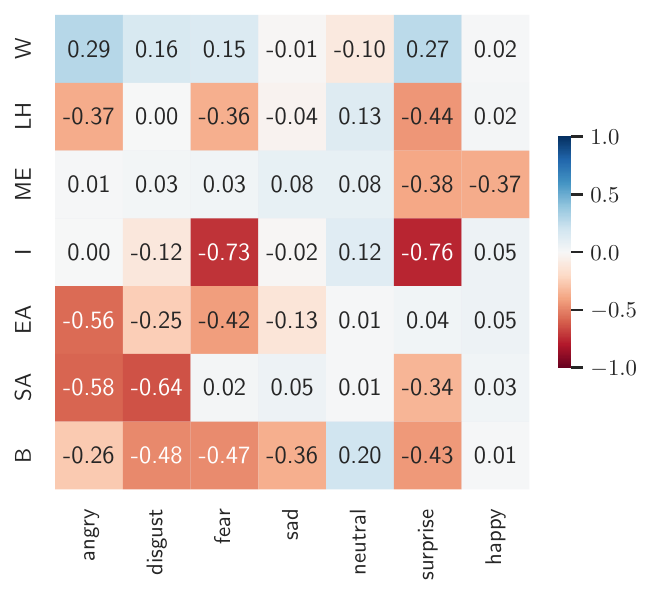}
        \includegraphics[width=.96\columnwidth]{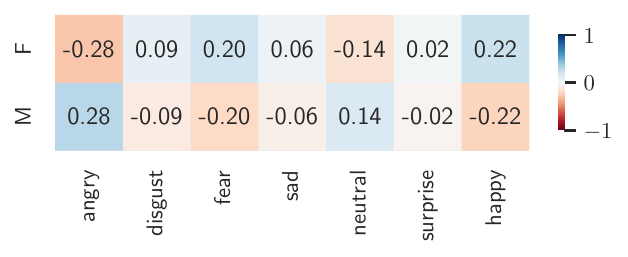}
    \end{minipage}
    \caption{Local stereotypical bias metrics for the FER+$^{\text{ITW}}$ dataset across the three demographic components. The first column corresponds to the raw number of samples in each combination of demographic group and target label, the second column corresponds to the NPMI metric and the third column to the Z metric. For the two metrics, higher absolute values indicate higher local stereotypical bias. Negative values indicate underrepresentation and positive values overrepresentation.}
    \label{figure:stereolocal}
\end{figure*}

\noindent\textbf{Overview}. In the case of local stereotypical bias the metrics report not a single value for a dataset and component combination, but a matrix, making impractical to include the full results here. Due to this, Figure~\ref{figure:stereolocal} shows an example of the application of local stereotypical bias metrics, in this case for the FER+ dataset, while the results for the rest of the datasets are included in the Supplementary Material, available online. To illustrate this case, we show the support of each subgroup (first column), in addition to the two stereotypical bias metrics, namely, NPMI (second column) and Z (third column). The support of the subgroups contains the full information needed to identify the local biases, but are hard to interpret manually, especially in cases where the target label, the demographic groups, or both of them are unbalanced or representationally biased, such as in this example. 

\vspace{0.3em}\noindent\textbf{Statistical agreement assessment}. Overall, we can intuitively observe a high correlation in the rank sorting of the biases detected by the two metrics, in this case when comparing the results within a single dataset and demographic component. We can again confirm this correlation by using the Spearman's $\rho$. In particular, for each dataset and demographic component we compute a NPMI and a Z matrix and compute the $\rho$ agreement score between them. The final averaged value is $\rho = 0.96 \pm 0.02$, indicating a strong correlation between the two metrics.

\vspace{0.3em}\noindent\textbf{Interpretability}. Despite the similarity in the produced rankings, the range of values of both metrics is remarkably different. In particular, the chosen example dataset highlights the main shortcoming of the NPMI metric, an oversensitivity to missing subgroups, such as the 70+ age group in the disgust label. These underrepresentation biases obtain a NPMI of $-1$, but arguably similar biases, such as the small representation of the 70+ age group in the fear label, obtain only a NPMI of $-0.22$, scaling in an unintuitive way. In the Z metric, we can observe values of $-1$ and $-0.9$ for these same groups, closer to the natural intuition. Generally speaking, the observed values for Z are better distributed in the range, making them more interpretable than those of the NPMI.

\vspace{0.3em}\noindent\textbf{Recommended metrics.} From these results, we recommend the usage of Z to evaluate local stereotypical bias, as the biases detected are similar to those of the NPMI, but having a better interpretability. 

\subsection{Demographic bias in FER datasets}\label{ssec:bias_in_fer}

In this section, we focus on analyzing the FER case study, employing the information provided by the metrics applied in the previous sections to identify which biases are the most prevalent across the datasets, and which datasets are the least affected by them. For the analysis in this section we group the datasets according to the source of the data in each dataset (according to the classification presented in Section~\ref{ssec:datasets}), as this is one of the key factors determining the type of biases exhibited.

\begin{figure*}[hbt]
    \centering
    \graphicspath{{images/fer_summary/}}
    \includegraphics[width=\textwidth]{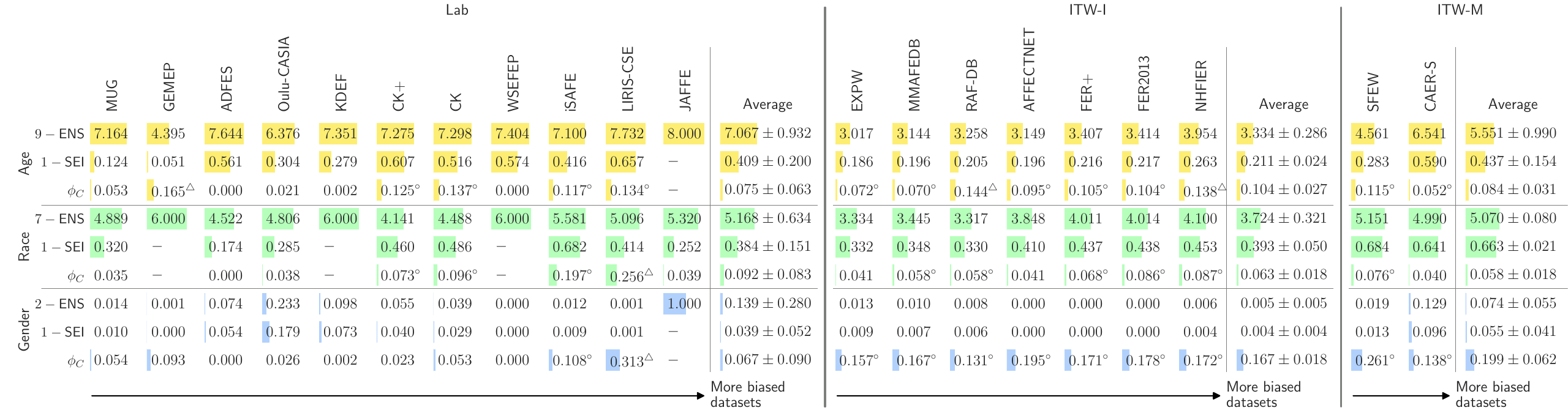}
    \caption{Summary of demographic bias in FER datasets, using the selected metrics. Higher values in any metric indicate a higher amount of bias. The representational bias metrics are shown in their bias formulation, in the case of ENS by subtracting it from the number of groups in each demographic component, and in the case of SEI from 1, its theoretical maximum. The datasets in each group are sorted by the average of the normalized metrics.}
    \label{figure:fer_summary}
\end{figure*}

Figure~\ref{figure:fer_summary} summarizes the findings in Figure~\ref{figure:rep_bias_detail} and Figure~\ref{figure:stereo_bias_detail}, according to the suggested metric selection of the previous sections. In particular, for representational bias we show the ENS and SEI to characterize representational bias, and $\phi_C$ to characterize stereotypical bias. In the case of ENS and SEI, we convert them to their bias variant, to facilitate their interpretation. In the case of ENS we use $|\text{G}|-\text{ENS}$, where $|\text{G}|$ is the number of groups in the demographic component, with the meaning of equivalent number of \textit{unrepresented} groups. In the case of SEI, we use $1 - \text{SEI}$, with the meaning of \textit{unevenness} between represented groups. The results are split in three sections, according to the data source of each dataset (LAB, ITW-I and ITW-M).

\vspace{0.3em}\noindent\textbf{Representational bias.} Overall, we observe the lowest representational bias values in the ITW-I datasets, followed by the ITW-M and finally the LAB datasets. Gender is the least biased component across the three groups, followed by age, and race exhibits the highest bias. We observe almost no representational bias in the gender component for all datasets, with most having a $|\text{G}|-\text{ENS}$ close to $0$. The exception to this is JAFFE, a laboratory-gathered dataset taken from a small sample of Japanese women that was never intended to be used as a general dataset for ML training~\cite{Lyons2021}. In the age and race components, the biases are more generalized, with the LAB datasets exhibiting the highest representational bias, while the ITW-M and ITW-I datasets seem to be less biased. Despite this, the ITW-M appear to be closer to the representation profiles of the LAB datasets than to those of the ITW-I datasets. Globally, EXPW, MMAFEDB and RAF-DB are the least representationally biased datasets, with low $|\text{G}|-\text{ENS}$ in the age ($\leq 3.258$) and the race ($\leq 3.445$) components. For the evenness aspect of representational bias, we can observe the results of $1-\text{SEI}$, which is relatively low and homogeneous across the ITW-I datasets. The two ITW-M datasets are less even across the represented groups than either the ITW-I and LAB datasets. Finally, the LAB datasets have a large and less coherent range of evenness values, showing a larger variety of demographic profiles between them.

\vspace{0.3em}\noindent\textbf{Global stereotypical bias.} Overall, stereotypical bias seems to be present at a lower rate than representational bias for the three groups, with only weak and medium bias found in some cases according to the $\phi_C$ metric thresholds. The least biased group is the LAB group in two of the components (gender and age), while the ITW-M is the least biased in the remaining one (race). As expected, this type of bias is almost absent from most LAB datasets, as they usually take samples for all the target classes for each subject. Despite this, some LAB datasets, such as LIRIS-CSE, GEMEP, iSAFE, CK, and CK+, do not follow this rule. These datasets only include certain classes for each subject, and in these cases they exhibit as much stereotypical bias as the ITW-I datasets. This effect is present mostly in the age and race components, while in the gender category these LAB datasets still exhibit lower bias scores than the ITW-I ones. ITW-I datasets have an overall higher stereotypical bias, with most of them showing weak or medium bias in the three demographic components, especially for the gender component. From the ITW-I datasets, the least stereotypically biased are EXPW and MMAFEDB. The two ITW-M, namely, SFEW and CAER-S, show very different stereotypical bias profiles, with SFEW having higher stereotypical bias scores in the three demographic components, while CAER-S is unbiased in the race component and less biased in the two other components compared to SFEW. 

\begin{figure}[hbt]
    \centering
    \begin{minipage}{0.29\textwidth}
        \includegraphics[width=\columnwidth]{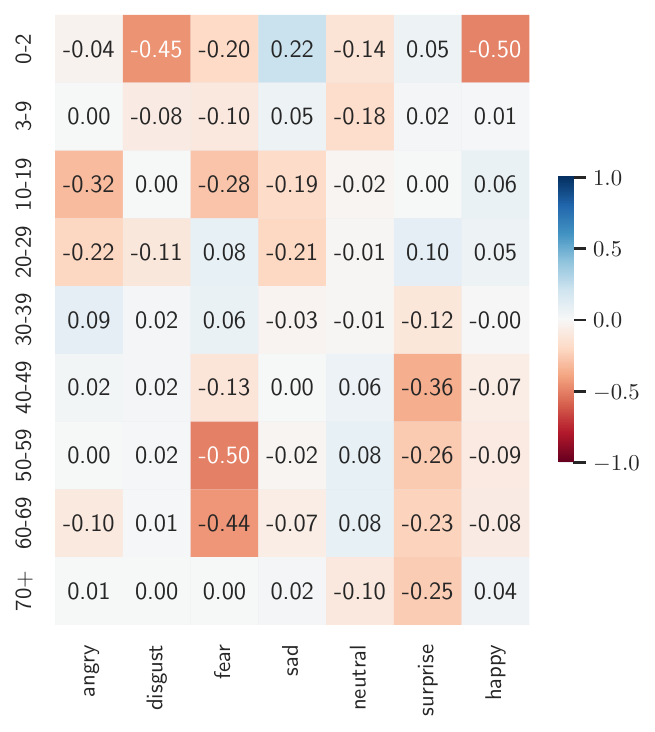}
        \includegraphics[width=\columnwidth]{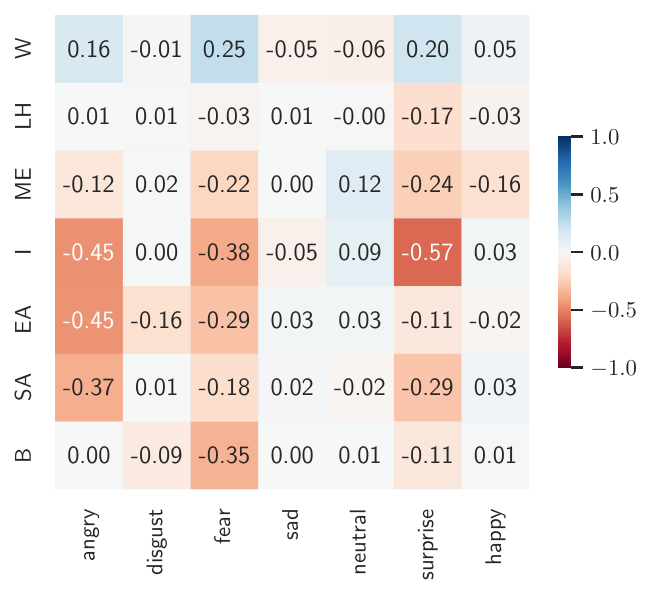}
        \includegraphics[width=.965\columnwidth]{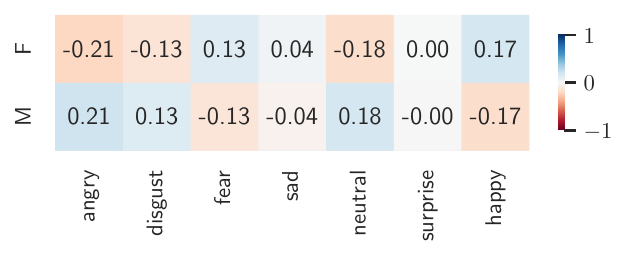}
    \end{minipage}
    \caption{Local stereotypical bias according to Z for the EXPW$^{\text{ITW}}$ dataset in each of the demographic components.}
    \label{figure:stereolocal_expw}
\end{figure}

\vspace{0.3em}\noindent\textbf{Local stereotypical bias.} According to these results, we can highlight EXPW as one of the least biased datasets overall. For this reason, in Figure~\ref{figure:stereolocal_expw} we measure its local stereotypical bias using Z. The highest biases are found in the age and race components, where six subgroups have large underrepresentations ($\text{Z}\leq -0.45$), namely the $0-2$ age group in the disgust and happy labels, the $50-59$ group in the fear label, the east asian and indian groups in the angry label, and the indian group in the surprise label. The angry, fear, and surprise labels exhibit the highest overall local stereotypical biases, especially in the race and age components.

\vspace{0.3em}\noindent\textbf{FER dataset selection.} With this information, currently no completely unbiased dataset is found for FER. Although ITW datasets such as RAF-DB, MMAFEDB, and EXP have relatively low representational bias and high evenness, their stereotypical bias results demand caution when using them as a sole source of training data. From these, EXPW can be highlighted as the least stereotypically biased dataset of the ITW category, and having a relatively large number of samples at $91,793$ images, it can be generally recommended. When using this dataset, precautions should be taken to evaluate the potential impact of stereotypical bias, especially in the angry, fear, and surprise labels in the race component.

\section{Conclusion and Future Work}\label{sec:conclusion}

In this work, we have proposed a taxonomy of metrics applicable to some of the main types of demographic biases found in datasets, namely, representational and stereotypical bias. We have incorporated into our review both metrics previously employed for this purpose and new proposals adapted from other fields, such as information theory and ecology. In particular, we have shown how metrics intended for species diversity in ecosystems can be also used to measure representational bias in datasets. After presenting these metrics, we have employed FER as a case of study, comparing the biases present in $20$ datasets for this task to evaluate the behavior of the metrics. With this information, we are able to highlight the most interpretable metrics for each subtype of bias, while avoiding redundant metrics that do not offer additional information.

Regarding representational bias, defined as an unequal representation of different demographic groups, we have found a relatively high experimental agreement between the different metrics available. Despite the diverse theoretical basis and implementation details, we conclude that the joint usage of a combined metric, where we suggest the Effective Number of Species (ENS) for its interpretability, and an evenness metric, such as the Shannon Evenness Index (SEI), selected for the same reason, are generally sufficient to characterize representational bias.

Regarding stereotypical bias, an undesired association between a demographic component and the target label, we find no significant difference between the metrics available. For interpretability reasons, we recommend the usage of Cramer's V ($\phi_C$) when measuring stereotypical bias in a dataset. Additionally, for the metrics used to measure stereotypical bias in a local way, that is, for a specific subgroup defined by both a demographic group and a target label, we suggest using Ducher's Z, as an interpretable and informative metric.

As a case study, we have applied these metrics to a collection of twenty FER datasets. We find evidence of representational bias in most of the datasets, especially those taken in laboratory conditions, as the low subject number and collection conditions lead to constrained demographic profiles. The datasets taken in the wild, especially those from internet searches, exhibit lower representational bias, but at the cost of higher stereotypical bias. Overall, we find that the EXPW dataset exhibits the lowest representational bias, with relatively low stereotypical bias. Furthermore, we apply a local stereotypical bias metric to identify the specific stereotypical biases that could be of concern, and find that special considerations should be taken when analyzing the angry, fear and surprise labels, as they are racially biased.

For future work, we can note that although this work has focused on representational and stereotypical bias, dataset demographic bias can manifest in many other ways~\cite{Fabbrizzi2022}, such as image quality, image context and label quality, and further research is still needed in the way these other manifestations can be measured. Additionally, we have only considered both demographic components and target variables as nominal variables, but our research could be extended to continuous and ordinal demographic components and regression problems. Furthermore, we have focused on stereotypical bias as an association between demographic components and target classes, although the association between several demographic components also poses potential risks and can be measured with the same metrics. Currently, there has been no research on the potential impact of biases in this regard.

Another limitation that could be improved in future work is the usage of demographic labels derived from a demographic relabeling model, FairFace. As these labels can be inaccurate, new datasets that include demographic information, or new models capable of more accurate demographic predictions, could support more robust bias analysis. 

The dataset bias analysis found here is supported by previous work~\cite{Dominguez-Catena2022,Dominguez-Catena2023} that has shown bias transference to the final trained model in the specific context of FER. Nonetheless, further work is still required to comprehend the implications and reach of this bias transference in different problems and contexts.

In this work, we have focused on how to measure demographic bias, but there is still work to be done on the usage of the reviewed metrics to study the transference of bias from the dataset to the model, as a way to improve bias mitigation strategies. To this end, we suggest the generation of intentionally biased synthetic datasets derived from real datasets as a general application-agnostic framework to evaluate both the limits of these demographic bias metrics and potential mitigation strategies. In this sense, and to inspire future works, in the Supplementary Material we provide the results of a series of experiments showcasing this methodology and showing how different types of dataset bias, measured with our proposed metrics, propagate in different ways to the final model predictions.

Finally, it is crucial to understand any bias results not only as a statistical discovery but as a potential harm to real people. In this sense, more work is needed on the psychological and sociological impact of potentially biased systems in the final applications.

\FloatBarrier

\clearpage

\begin{appendices}

In the following supplementary material we include additional results that complement the main paper. Section~\ref{apB} presents a series of experiments on the propagation of dataset biases into the trained models, showing how different types of biases propagate differently to the trained model predictions. Afterwards, in Section~\ref{apA} we provide the full contingency tables resulting from the demographic analysis of the twenty FER datasets, together with the NPMI and Z matrices for local stereotypical bias detection. The global stereotypical bias and representational bias results are available in the main paper.


\section{Downstream bias propagation}\label{apB}


    In this section, we present a series of experiments to show the relationship between dataset bias and downstream bias propagated to the model. In these experiments, our objective is to study how different bias types, measured according to our proposed taxonomy and metrics, affect the final model. The ideal case would be to find straightforward linear relationships between the bias in the dataset and the perturbation on the model, but our results suggest that reality is much more complex.

    In the following, we provide details on the four experiments carried out. These experiments are designed to evaluate different potential bias modalities, from the more common representational bias case to several stereotypical bias scenarios. We want to show how bias metrics can summarize existing bias in the source datasets and how those biases are transferred to the model. To do so, we focus on a single source dataset and generate biased subsets of it, analyzing how this affects the final performance of the models.

    The methodology of the four experiments is similar, only changing the type of induced bias between them. The experimental details are as follows:

    \begin{itemize}
        \item \textbf{Data source.} As a source dataset, we selected Affectnet~\cite{Mollahosseini2019}, one of the largest datasets for FER. The size of this dataset enables us to create strongly biased subsets that are still large enough to train performant models.
        \item \textbf{Demographic component.} For simplicity, we focus only on the gender demographic component, as predicted by FairFace in the experiments in the article (Section IV-C). The gender component is binary coded, with both gender groups roughly equally represented in Affectnet. This maximizes the usable portion of the dataset when generating biased sampled datasets and keeps the analysis of the model output manageable.
        \item \textbf{Bias proportion.} The specific definition of bias changes between the four experiments, but in all cases it is defined with a number between $0.0$ and $1.0$, referring to the proportion of induced bias. We created datasets for bias amounts in this range in increments of $0.1$, for a total of $11$ datasets per experiment.
        \item \textbf{Dataset generation.} When extracting the subsets of Affectnet to be used as biased datasets, the same class proportions and total size are kept for each experiment, to guarantee the comparability of the results. The size of each class is set to the size of the least represented demographic group in that class. This guarantees that, when that group is favored by the induced bias, there are enough samples to fill the entire class if necessary. Additionally, the resulting class distribution is more realistic, closely following the real distribution of Affectnet and other datasets, which rarely have equal class sizes.
        \item \textbf{Bias metrics.} Dataset bias in the generated datasets is evaluated using the set of metrics recommended in Sections 5.1 and 5.2 of the main paper, namely, Effective Number of Species (ENS), which summarizes the representational bias of the dataset, Shannon Evenness Index (SEI), which measures the evenness of the groups represented, and Cramer's V ($\phi_C$), which measures stereotypical bias.
        \item \textbf{Model and model training.} For each dataset a pre-trained ResNet50 model~\cite{He2015} is trained for $20$ epochs, following a 1cycle train policy (as described in \cite{Smith2018}) with a maximum learning rate of $1e^{-4}$. 
        \item \textbf{Downstream bias evaluation.} After training, each model is evaluated on the entire Affectnet test partition, and the recall for each class and gender group is recorded individually. These recalls are a robust way of measuring the model predictions despite biases in the test partition. The process is repeated three times, and the reported recalls are the average of the tree repetitions of each configuration, as a way of stabilizing the stochastic nature of the dataset subsampling and the training process.
    \end{itemize}
    
    The four experiments vary depending on the analyzed type of bias, as follows.
    \begin{itemize}
        \item \textit{Experiment 1. Representational bias.} In this experiment, we investigate the impact of a general underrepresentation of a gender group in a dataset.
        \item \textit{Experiment 2. Single class stereotypical bias.} In this experiment, we study how stereotypical bias affecting a single class propagates to the trained model. To simulate this situation, we modify the composition of one of the classes, the ``happy'' class. As it is the largest class in the dataset, it allows us to modify the distribution of samples in greater absolute numbers.
        \item \textit{Experiment 3. Opposite multiclass stereotypical bias.} When studying stereotypical bias, in the paper we encountered that most classes were simultaneously biased. In this experiment, we test whether stereotypical bias in two different classes interact with each other. To evaluate this situation, we modify two classes in the datasets, ``happy'' and ``angry'', one having the inverse gender ratios of the other. This particular experimental setup corresponds to the classical ``angry-men-happy-women'' bias that we have encountered in most public FER datasets gathered from Internet.
        \item \textit{Experiment 4. Aligned multiclass stereotypical bias.} Extending on the previous experiment, in this one we study how stereotypical biases in several classes can interact with each other. In this case, we study the potential combination of stereotypical biases when several classes are biased in the same direction, that is, if the same demographic group is favored or disfavored by several classes. This type of interaction means the underrepresentation of a group not in a single class, but in several. From the perspective of the model, it could be similar to the overrepresentation of that same group in the unbiased classes, potentially spreading the bias to other classes. To test this, we modify two classes, ``happy'' and ``angry'', both sharing the same gender ratios.
    \end{itemize}

    \subsection{Experiment 1: Representational bias}

    \textbf{Objective.} The main objective of this experiment is to investigate how gender representational bias in the source dataset affects the performance of the trained model model.

    \textbf{Biased datasets.} For the generation of the representationally biased datasets, the amount of induced bias is defined as the proportion of female subjects in the resulting dataset. Table~\ref{tab:exp1} shows the contingency tables of three of these datasets. In particular, the top section corresponds to a data set without female subjects (proportion $0.0$), the middle section corresponds to an unbiased dataset with equal representation (proportion $0.5$), and the bottom section to one with only female subjects (proportion $1.0$).

    \begin{table*}[ht]
        \centering
        {\bfseries\strut Dataset 1}\\
        \begin{tabular}{rlrrrrrrrr}
        \toprule
        Proportion & Gender & Angry & Disgust & Fear & Happy & Neutral & Sad & Surprise & Total \\
        \midrule
        \multirow{2}*{0.0} & Female & 0 & 0 & 0 & 0 & 0 & 0 & 0 & 0 \\
         & Male & 6,962 & 1,733 & 3,117 & 54,241 & 33,708 & 11,175 & 6,439 & 117,375 \\
        \bottomrule
        \end{tabular}
        
        \vspace{1em}
        
        {\bfseries\strut Dataset 2}\\
        \begin{tabular}{rlrrrrrrrr}
        \toprule
        Proportion & Gender & Angry & Disgust & Fear & Happy & Neutral & Sad & Surprise & Total \\
        \midrule
        \multirow{2}*{0.5} & Female & 3,481 & 866 & 1,558 & 27,120 & 16,854 & 5,587 & 3,219 & 58,685 \\
         & Male & 3,481 & 866 & 1,558 & 27,120 & 16,854 & 5,587 & 3,219 & 58,685 \\
        \bottomrule
        \end{tabular}
        
        \vspace{1em}
        
        {\bfseries\strut Dataset 3}\\
        \begin{tabular}{rlrrrrrrrr}
        \toprule
        Proportion & Gender & Angry & Disgust & Fear & Happy & Neutral & Sad & Surprise & Total \\
        \midrule
        \multirow{2}*{1.0} & Female & 6,962 & 1,733 & 3,117 & 54,241 & 33,708 & 11,175 & 6,439 & 117,375 \\
         & Male & 0 & 0 & 0 & 0 & 0 & 0 & 0 & 0 \\
        \bottomrule
        \end{tabular}

        \vspace{1em}
        
        \caption{Contingency tables for three datasets with varying degrees of representational bias.}
        \label{tab:exp1}
    \end{table*}
    
\begin{figure*}[ht]
        \centering
        \subfigure[]{\includegraphics[width=0.32\linewidth]{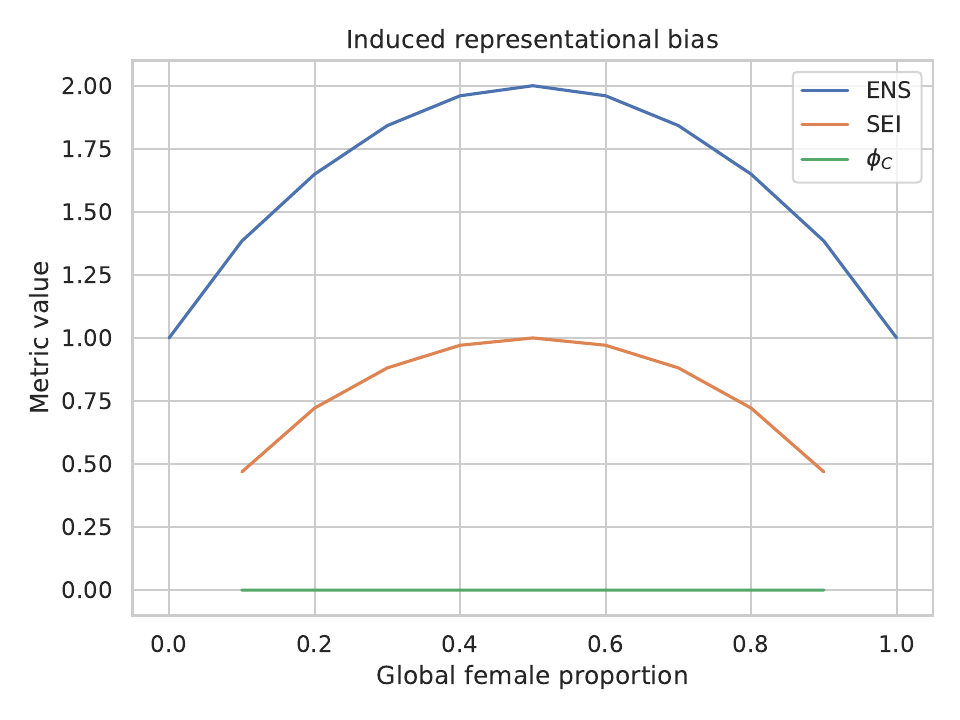}
        \label{fig:exp1a}}
        \subfigure[]{\includegraphics[width=0.32\linewidth]{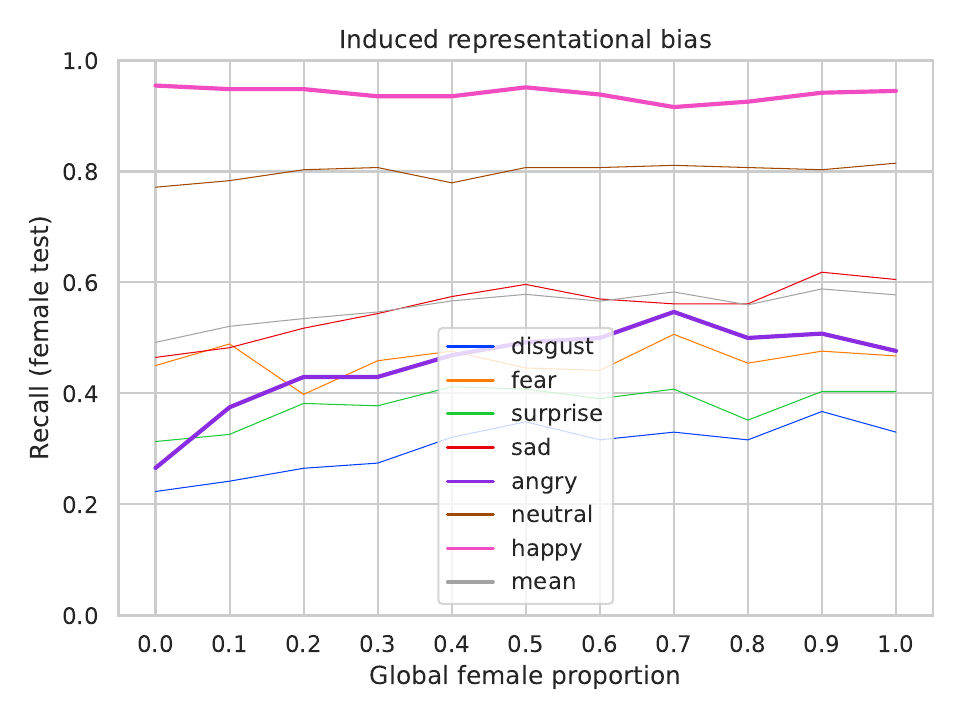}
        \label{fig:exp1b}}
        \subfigure[]{\includegraphics[width=0.32\linewidth]{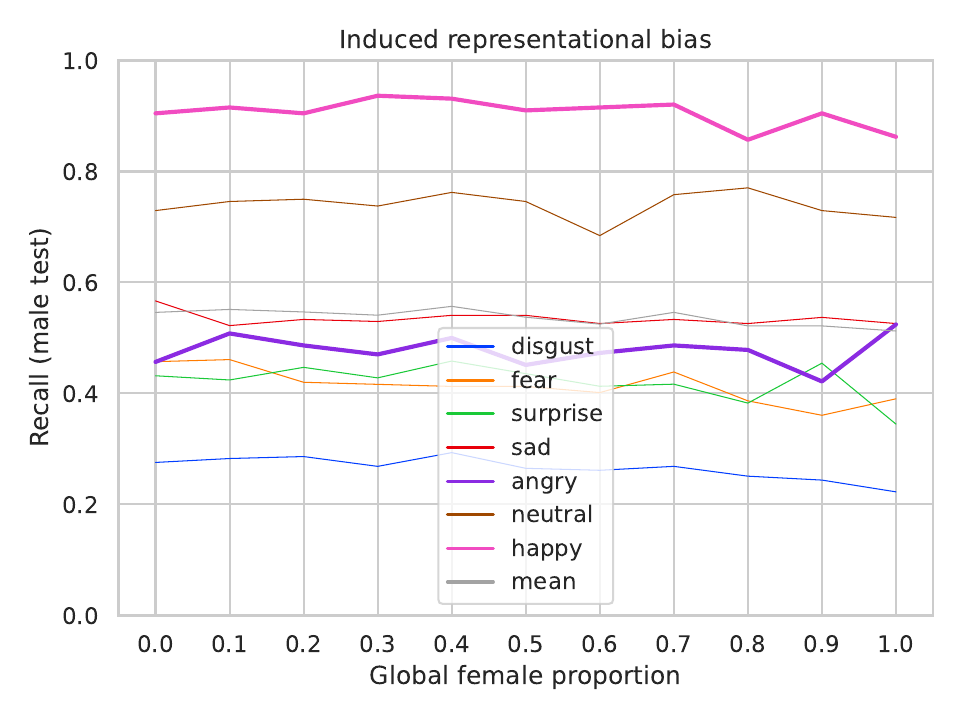}
        \label{fig:exp1c}}
        \caption{(a) Bias measures of each representationally biased dataset. (b) Recall per class for the female group. (c) Recall per class for the male group. For all three, in the horizontal axis, amount of induced bias.}
        \label{fig:exp1}
    \end{figure*}

    \textbf{Dataset bias metrics.} In Figure~\ref{fig:exp1a} we can observe the results of applying three of the dataset bias metrics that are proposed in the paper. In particular, the blue line represents the Effective Number of Species (ENS), which summarizes the representational bias of the dataset. As expected, this line curves to both sides (proportions 0 and 1), which have a single gender group represented, and thus an ENS of exactly 1. The proportion $0.5$, corresponding to a balanced dataset, has an ENS of 2, as it represents two groups of equal sizes. The orange line, representing the Shannon's Evenness Index (SEI), shows the measured evenness of the datasets independently of the total number of groups represented. It also gets maximized at the center (proportion $0.5$ with a SEI of exactly $1$, perfectly even representation), and decreases towards the sides, getting undefined when only one group is represented. Finally, the green line corresponds to stereotypical bias, as measured by the Cramer's V ($\phi_C$). In these cases, as all the datasets have the same proportion of female to male subjects per class, no stereotypical bias is identified and $\phi_C$ equals $0$ across the range.

    \textbf{Model performance.} Figures~\ref{fig:exp1b} and~\ref{fig:exp1c} show the recall of the trained models for each class, for the female and male genders, respectively. The effect of representational bias appears relatively weak, with only the extreme cases (fraction $0.0$ and fraction $1.0$) showing a mild perturbation of the recalls. In general, the strongest effect is for the angry class in the female group, where a $20\%$ increase in recall can be observed between the two extremes of bias (fraction $0.0$ to fraction $1.0$). The rest of the classes have variations up to $10\%$, most of them in the cases of extreme bias, especially in the female group.

    \textbf{Conclusions.} In general, the effect of representational bias between the two genders seems weaker than expected, with the models being able to properly generalize even when one of the genders disappears from the dataset with a slight drop of accuracy. It is important to notice that this is proven only for this experimental setup and for the gender demographic component. Other components, such as age, have been shown to be susceptible to this type of bias~\cite{Kim2021}, and the same component in other problems~\cite{Buolamwini2018}.

    \subsection{Experiment 2: Single class stereotypical bias}

    \textbf{Objective.} In this experiment, we assess the impact of stereotypical bias in its most trivial form, when a single class is affected. To simulate this case, we focus on the happy class (the majority one), generating datasets that have different proportions of female subjects in it.
    
    \textbf{Biased datasets.} We induce stereotypical bias for this experiment by modifying the number of samples belonging to the happy class, as it is the class with a larger number of samples. This enables larger amounts of bias to be tested. In this case, the amount of induced bias is measured by the proportion of female subjects in the happy class. Table~\ref{tab:exp2} shows three of these stereotypically biased datasets, with proportions $0.0$, $0.5$, and $1.0$. Note that in these datasets only the happy class is perturbed, keeping a balanced representation for the rest of the classes in the dataset.

    \begin{table*}[ht]
        \centering
        {\bfseries\strut Dataset 1}\\
        \begin{tabular}{rlrrrrrrrr}
        \toprule
        Proportion & Gender & Angry & Disgust & Fear & Happy & Neutral & Sad & Surprise & Total \\
        \midrule
        \multirow{2}*{0.0} & Female & 3,481 & 866 & 1,558 & 0 & 16,854 & 5,587 & 3,219 & 31,565 \\
         & Male & 3,481 & 866 & 1,558 & 54,241 & 16,854 & 5,587 & 3,219 & 85,806 \\
        \bottomrule
        \end{tabular}
        
        \vspace{1em}
        
        {\bfseries\strut Dataset 2}\\
        \begin{tabular}{rlrrrrrrrr}
        \toprule
        Proportion & Gender & Angry & Disgust & Fear & Happy & Neutral & Sad & Surprise & Total \\
        \midrule
        \multirow{2}*{0.5} & Female & 3,481 & 866 & 1,558 & 27,120 & 16,854 & 5,587 & 3,219 & 58,685 \\
         & Male & 3,481 & 866 & 1,558 & 27,120 & 16,854 & 5,587 & 3,219 & 58,685 \\
        \bottomrule
        \end{tabular}
        
        \vspace{1em}
        
        {\bfseries\strut Dataset 3}\\
        \begin{tabular}{rlrrrrrrrr}
        \toprule
        Proportion & Gender & Angry & Disgust & Fear & Happy & Neutral & Sad & Surprise & Total \\
        \midrule
        \multirow{2}*{1.0} & Female & 3,481 & 866 & 1,558 & 54,241 & 16,854 & 5,587 & 3,219 & 85,806 \\
         & Male & 3,481 & 866 & 1,558 & 0 & 16,854 & 5,587 & 3,219 & 31,565 \\
        \bottomrule
        \end{tabular}
        
        \vspace{1em}
        
        \caption{Contingency tables for three datasets with varying degrees of stereotypical bias in the happy class.}
        \label{tab:exp2}
    \end{table*}

\begin{figure*}[ht]
        \centering
        \subfigure[]{\includegraphics[width=0.32\linewidth]{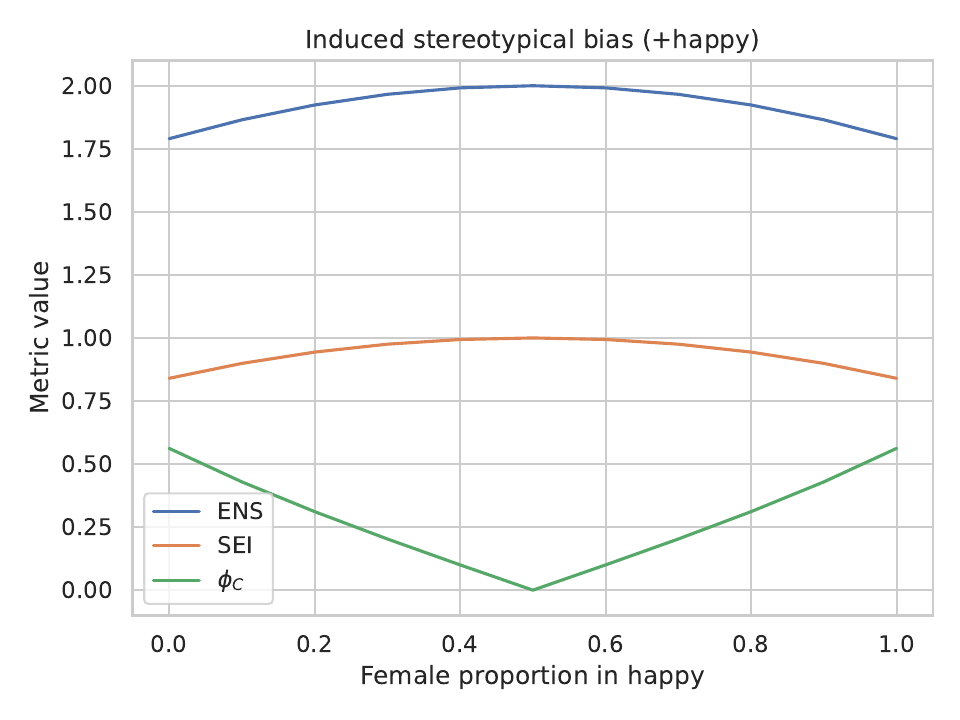}
        \label{fig:exp2a}}
        \subfigure[]{\includegraphics[width=0.32\linewidth]{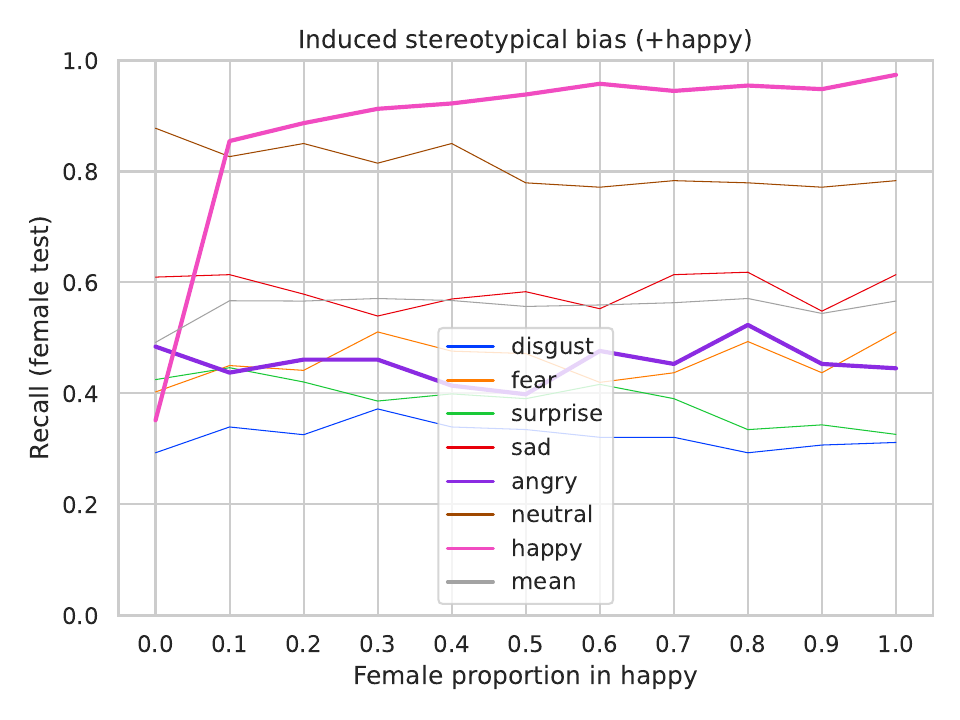}
        \label{fig:exp2b}}
        \subfigure[]{\includegraphics[width=0.32\linewidth]{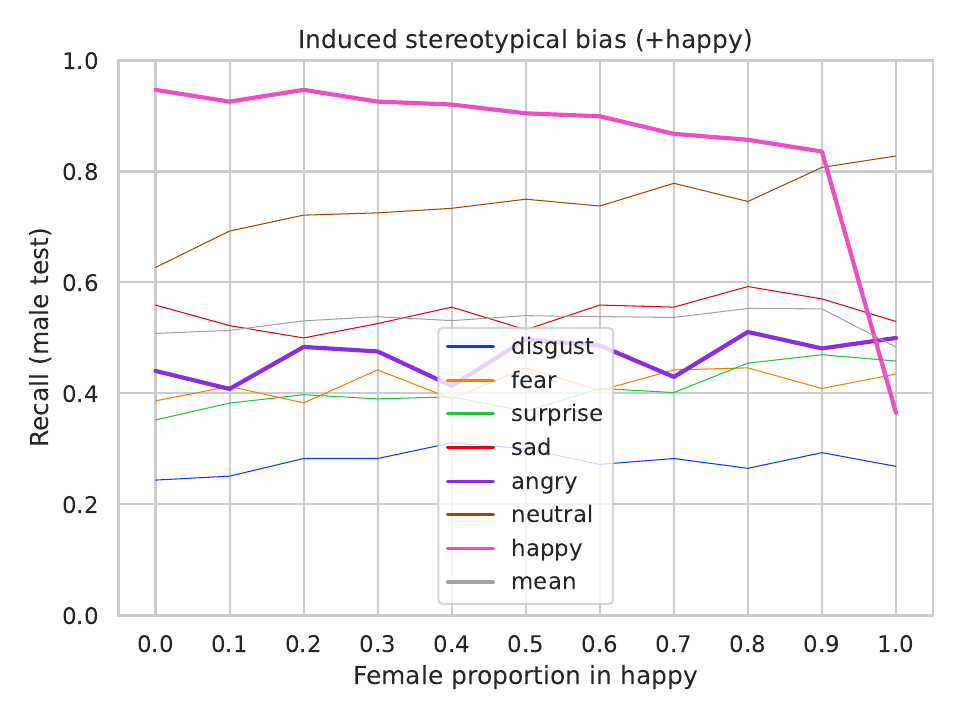}
        \label{fig:exp2c}}
        \caption{(a) Bias measures of each dataset stereotypically biased in the happy class. (b) Recall per class for the female group. (c) Recall per class for the male group. For all three, in the horizontal axis, amount of induced bias.}
        \label{fig:exp2}
    \end{figure*}

    \textbf{Dataset bias metrics.} In Figure~\ref{fig:exp2a} we measure the bias induced in the dataset with the proposed metrics. The ENS and SEI, which characterize representational bias and evenness, respectively, show a relatively weak variation between the different bias levels. Unfortunately, the only way to generate stereotypical bias without generating some representational bias as a side effect is to modify the proportion of female subjects in the rest of the classes, making it impossible to measure the effect of single-class stereotypical bias. Regarding the stereotypical bias metric $\phi_C$, plotted in green, a strong variation can be observed. As expected, a $0.0$ is obtained for the balanced dataset with fraction $0.5$, whereas it increases to more than $0.5$ for the  most biased proportions ($0.0$ and $1.0$).

    \textbf{Model performance.} In Figures~\ref{fig:exp2b} and \ref{fig:exp2c} we can observe how these perturbations affect the final trained models with respect to the female and male groups, respectively. The happy class, in particular, shows a strong variation in the most biased datasets (proportions $0.0$ and $1.0$). For the female gender group, there is a strong drop in recall for proportion $0.0$, that is, when there are no female examples in the happy class. A recall under $0.4$ is observed for the bias fraction $0.0$, whereas a recall over $0.85$ is observed for the rest of the bias fractions ($\geq 0.1$). The male gender group has a similar drop for the bias fraction $1.0$, corresponding to a lack of male representation in the happy class. In the female and male groups, the recall of the other classes does not appear to be significantly affected, with the strongest variation in the male group and the neutral class, with a recall drop around $0.2$, inversely correlated with the drop in the happy class. This can suggest a transference of predictions from one class to the other as the bias increases.

    \textbf{Conclusions.} The results show that for this experimental setup single-class stereotypical bias has a strong but localized effect, without significantly affecting the non-perturbed classes. Additionally, the experiments show the non-linearity of the effect, which in this particularly case is stronger in the extreme cases, when the underrepresented classes are in proportions under $0.1$, with a weaker influence on the rest of the range. In both genders, as the proportion of that gender diminishes, so does the recall of the affected class. It is important to notice that the induced dataset perturbation of the happy class in this experiment is identical to that in Experiment 1, with the difference that for this experiment the rest of the classes are kept balanced. This shows how, in the case of stereotypical bias, a smaller dataset perturbation can actually produce a larger bias in the model.

    \subsection{Experiment 3: Opposite multiclass stereotypical bias}

    \textbf{Objective.} In this experiment, we intend to observe how much multiple simultaneous stereotypical biases interact with each other. To do so, we bias two classes in opposite ways. This could produce a stronger impact than single-class stereotypical bias alone, since there is a potential risk of transference of predictions from the underrepresented class to the overrepresented one according to the perceived gender of the sample.

    \textbf{Biased datasets.} Following the same procedure as in the previous experiment (Experiment 2), we induce stereotypical bias in two classes (happy and angry) in opposite directions. This particular experimental setup corresponds to the classical ``angry-men-happy-women'' bias that we have encountered in most public FER datasets gathered from the Internet. As before, the amount of induced bias is measured by the proportion of female subjects in the happy class. The opposite (complementary) of this proportion is then used as the female proportion of the angry class. That is, the greater the female ratio for the happy class, the lower for the angry class.
    Table~\ref{tab:exp3} shows three of the generated datasets for this experiment, created with a bias parameter of $0.0$, $0.5$, and $1.0$, respectively. Observe that the rest of the classes are not biased in any way. However, since the two biased classes do not have exactly the same number of total subjects, some representational bias is still present.

    \begin{table*}[ht]
        \centering
        {\bfseries\strut Dataset 1}\\
        \begin{tabular}{rlrrrrrrrr}
        \toprule
        Proportion & Gender & Angry & Disgust & Fear & Happy & Neutral & Sad & Surprise & Total \\
        \midrule
        \multirow{2}*{0.0} & Female & 6,962 & 866 & 1,558 & 0 & 16,854 & 5,587 & 3,219 & 35,046 \\
         & Male & 0 & 866 & 1,558 & 54,241 & 16,854 & 5,587 & 3,219 & 82,325 \\
        \bottomrule
        \end{tabular}
        
        \vspace{1em}
        
        {\bfseries\strut Dataset 2}\\
        \begin{tabular}{rlrrrrrrrr}
        \toprule
        Proportion & Gender & Angry & Disgust & Fear & Happy & Neutral & Sad & Surprise & Total \\
        \midrule
        \multirow{2}*{0.5} & Female & 3,481 & 866 & 1,558 & 27,120 & 16,854 & 5,587 & 3,219 & 58,685 \\
         & Male & 3,481 & 866 & 1,558 & 27,120 & 16,854 & 5,587 & 3,219 & 58,685 \\
        \bottomrule
        \end{tabular}
        
        \vspace{1em}
        
        {\bfseries\strut Dataset 3}\\
        \begin{tabular}{rlrrrrrrrr}
        \toprule
        Proportion & Gender & Angry & Disgust & Fear & Happy & Neutral & Sad & Surprise & Total \\
        \midrule
        \multirow{2}*{1.0} & Female & 0 & 866 & 1,558 & 54,241 & 16,854 & 5,587 & 3,219 & 82,325 \\
         & Male & 6,962 & 866 & 1,558 & 0 & 16,854 & 5,587 & 3,219 & 35,046 \\
        \bottomrule
        \end{tabular}
        
        \vspace{1em}
        
        \caption{Contingency tables for three datasets with varying degrees of stereotypical bias in two classes in the opposite direction (happy and angry).}
        \label{tab:exp3}
    \end{table*} 

\begin{figure*}[ht]
        \centering
        \subfigure[]{\includegraphics[width=0.32\linewidth]{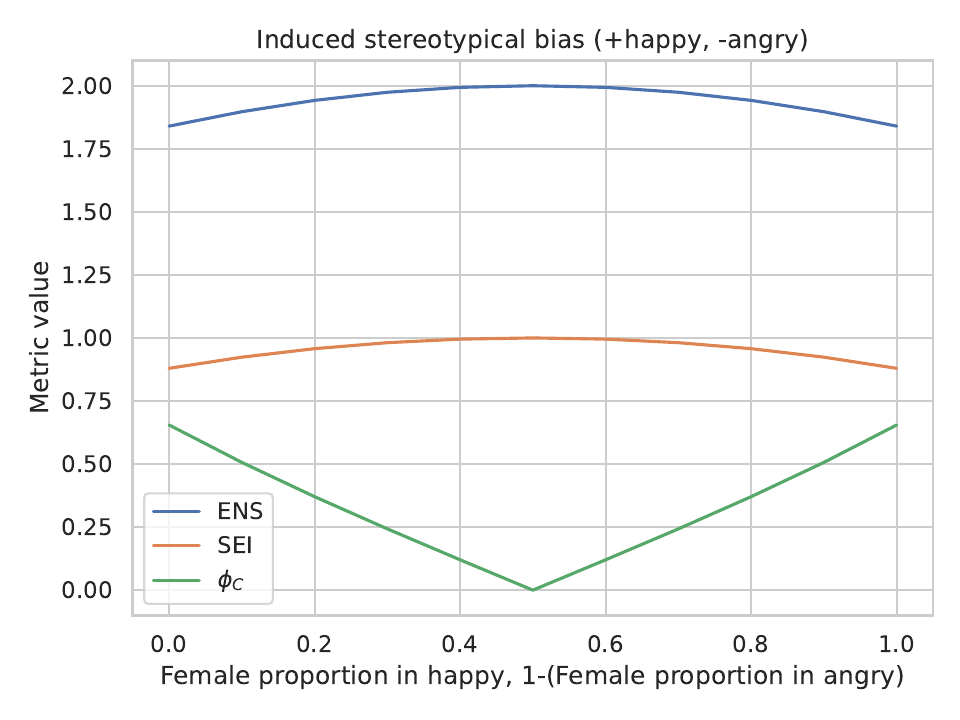}
        \label{fig:exp3a}}
        \subfigure[]{\includegraphics[width=0.32\linewidth]{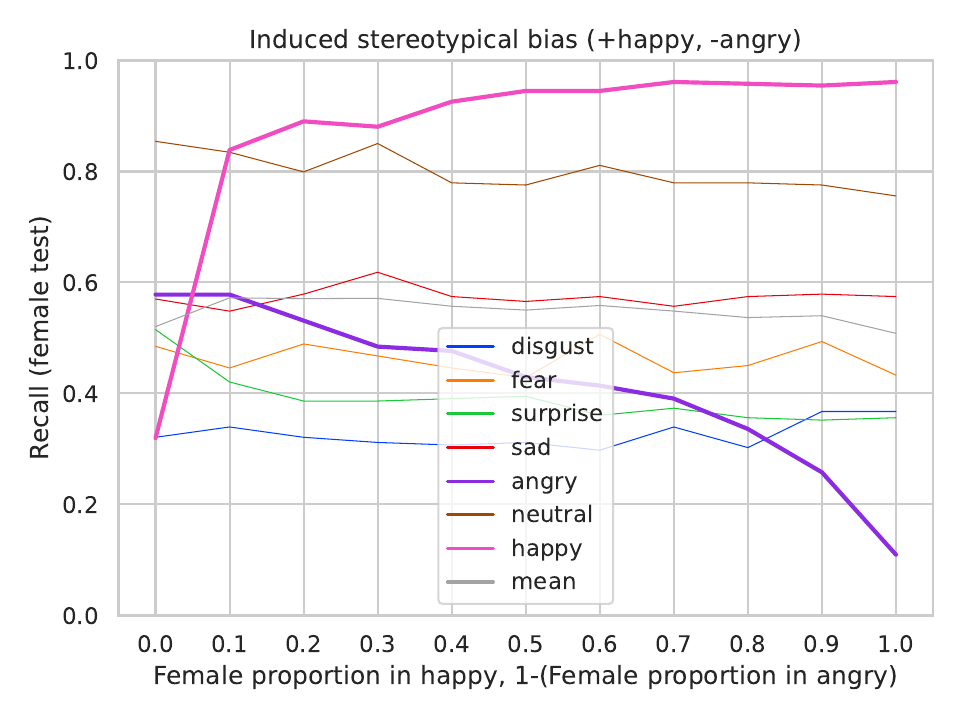}
        \label{fig:exp3b}}
        \subfigure[]{\includegraphics[width=0.32\linewidth]{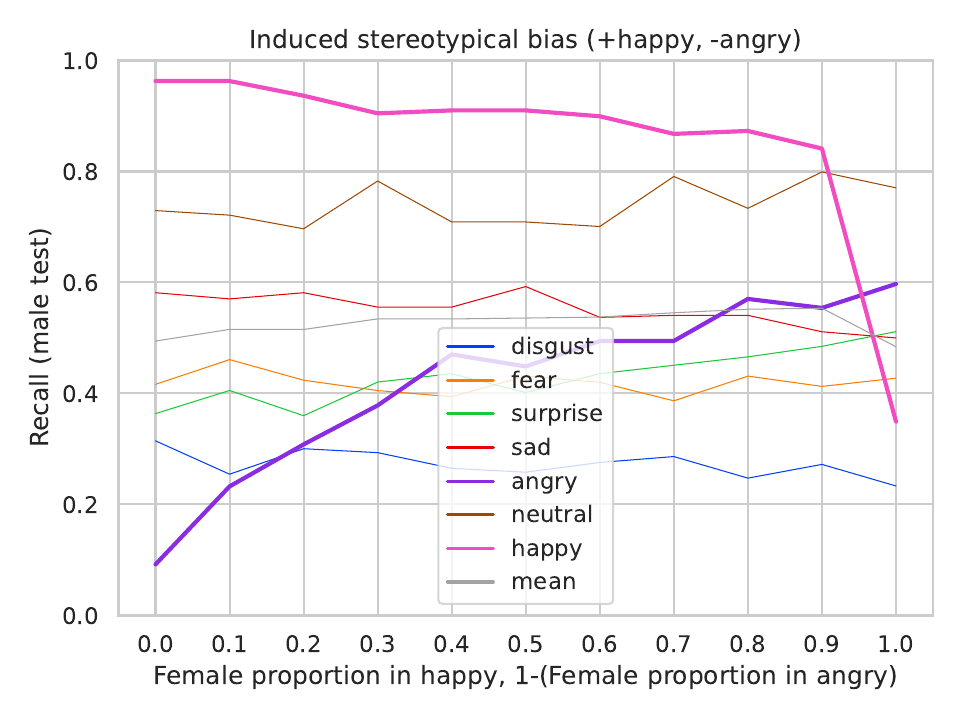}
        \label{fig:exp3c}}
        \caption{(a) Bias measures of each dataset stereotypically biased in happy and angry classes (opposite direction). (b) Recall per class for the female group. (c) Recall per class for the male group. For all three, in the horizontal axis, amount of induced bias.}
        \label{fig:exp3}
    \end{figure*}

    \textbf{Dataset bias metrics.} In Figure~\ref{fig:exp3a} a marginal amount of representational bias and unevenness can be observed in the datasets, as measured by ENS and SEI, respectively. This marginal bias is reflected in the slight decrease in both metrics when deviating from the balanced reference dataset ($0.5$ proportion). In this case, recall that this is due to the different total number of examples in each of the biased classes (happy and angry). In the same figure, the stereotypical bias metric $\phi_C$ shows a significant increase as the amount of bias approaches its limits at proportions $0.0$ and $1.0$, reaching over $0.6$ in both cases, showing that the induced stereotypical bias is even larger than the one induced in Experiment 2, where only the happy class was perturbed.

    \textbf{Model performance.} Figures~\ref{fig:exp3b} and \ref{fig:exp3c} represent the recalls per class of the female and male genders, respectively. Regarding the happy class (the pink lines), the effect of the induced stereotypical bias on the happy class is more widespread throughout the range of the bias perturbation than in Experiment 2, with a significantly larger effect in the extreme proportions ($0.0$ and $1.0$). With respect to the angry class, the recalls are more heavily affected, and the effect is more linear, occurring over the whole range of bias proportions. With a baseline recall of about $0.4$ for both gender groups (proportion $0.5$), both improve to around $0.6$ when the induced bias adds all the examples to that group (proportion $0.0$ for female and $1.0$ for male), but decrease to about $0.1$ when the induced bias goes the other way around, removing all examples for that group. For the rest of the classes, there is some weak effect, but generally with a recall variation below $0.1$.
    
    \textbf{Conclusions.} The impact of stereotypical bias varies greatly depending on the classes being affected. For example, for the happy class, the induced bias affects mainly in the extreme cases, while for the angry class, the induced bias affects throughout the whole range. In both perturbed classes, a diminished proportion of a gender group in train results in a loss in recall for that same group in test. However, the effect of the bias seems to be localized to the perturbed class, and there is no significant difference between the effect when only one class is perturbed and when two classes are perturbed in opposite ways. 
    
    \subsection{Experiment 4: Aligned multiclass stereotypical bias}

    \textbf{Objective.} In this experiment, we continue focusing on how multiple simultaneous stereotypical biases interact with each other. However, in this case we are interested in investigating whether biasing two classes in the same direction maximizes the probability of affecting the other classes.

    \textbf{Biased datasets.} We follow the same procedure aas in Experiment 3, but biasing both classes (happy and angry) in the same direction, meaning that if we bias in favor of the female gender group, we will have more females in both the happy and angry classes. The induced bias for this experiment is defined according to the proportion of the female group in both the happy and angry classes, with the rest of the classes maintaining the original balanced proportion ($0.5$).    Table~\ref{tab:exp4} shows the contingency tables of three of the  induced datasets with different amounts of bias ($0.0$, $0.5$ and $1.0$). Note also that as two classes are perturbed in the same direction, the side effect of representational bias is larger than in Experiments 2 and 3, but still far from the intentional representational bias in Experiment 1.

    \begin{table*}[ht]
        \centering
        {\bfseries\strut Dataset 1}\\
        \begin{tabular}{rlrrrrrrrr}
        \toprule
        Proportion & Gender & Angry & Disgust & Fear & Happy & Neutral & Sad & Surprise & Total \\
        \midrule
        \multirow{2}*{0.0} & Female & 0 & 866 & 1,558 & 0 & 16,854 & 5,587 & 3,219 & 28,084 \\
         & Male & 6,962 & 866 & 1,558 & 54,241 & 16,854 & 5,587 & 3,219 & 89,287 \\
        \bottomrule
        \end{tabular}
        
        \vspace{1em}
        
        {\bfseries\strut Dataset 2}\\
        \begin{tabular}{rlrrrrrrrr}
        \toprule
        Proportion & Gender & Angry & Disgust & Fear & Happy & Neutral & Sad & Surprise & Total \\
        \midrule
        \multirow{2}*{0.5} & Female & 3,481 & 866 & 1,558 & 27,120 & 16,854 & 5,587 & 3,219 & 58,685 \\
         & Male & 3,481 & 866 & 1,558 & 27,120 & 16,854 & 5,587 & 3,219 & 58,685 \\
        \bottomrule
        \end{tabular}
        
        \vspace{1em}
        
        {\bfseries\strut Dataset 3}\\
        \begin{tabular}{rlrrrrrrrr}
        \toprule
        Proportion & Gender & Angry & Disgust & Fear & Happy & Neutral & Sad & Surprise & Total \\
        \midrule
        \multirow{2}*{1.0} & Female & 6,962 & 866 & 1,558 & 54,241 & 16,854 & 5,587 & 3,219 & 89,287 \\
         & Male & 0 & 866 & 1,558 & 0 & 16,854 & 5,587 & 3,219 & 28,084 \\
        \bottomrule
        \end{tabular}

        \vspace{1em}
        
        \caption{Contingency tables for three datasets with varying degrees of stereotypical bias in two classes in the same direction (happy and angry).}
        \label{tab:exp4}
    \end{table*}

\begin{figure*}[ht]
        \centering
        \subfigure[]{\includegraphics[width=0.32\linewidth]{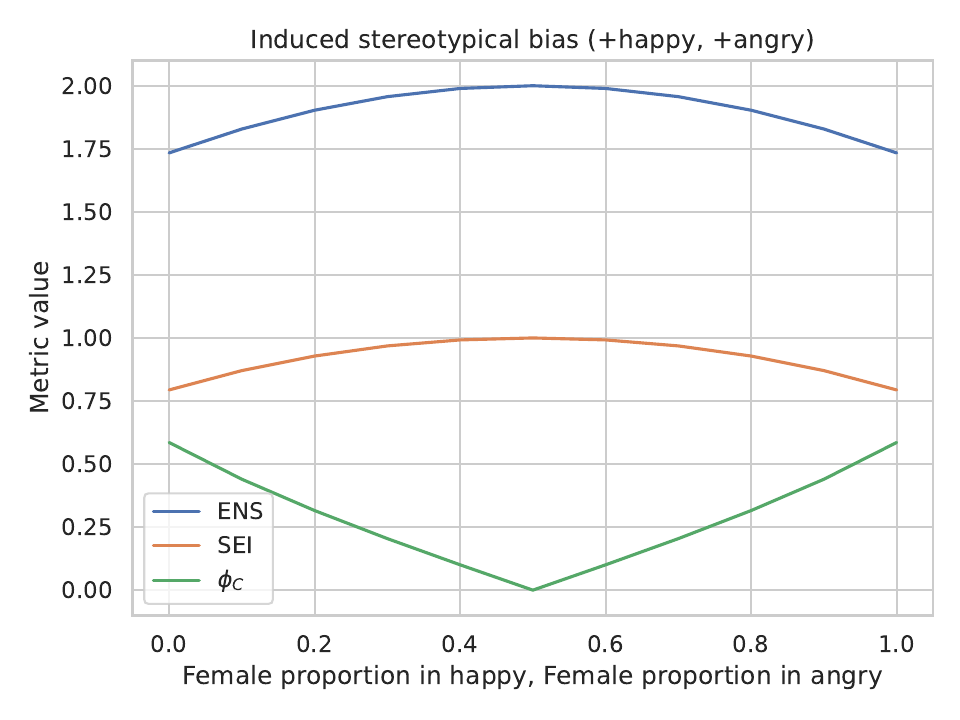}
        \label{fig:exp4a}}
        \subfigure[]{\includegraphics[width=0.32\linewidth]{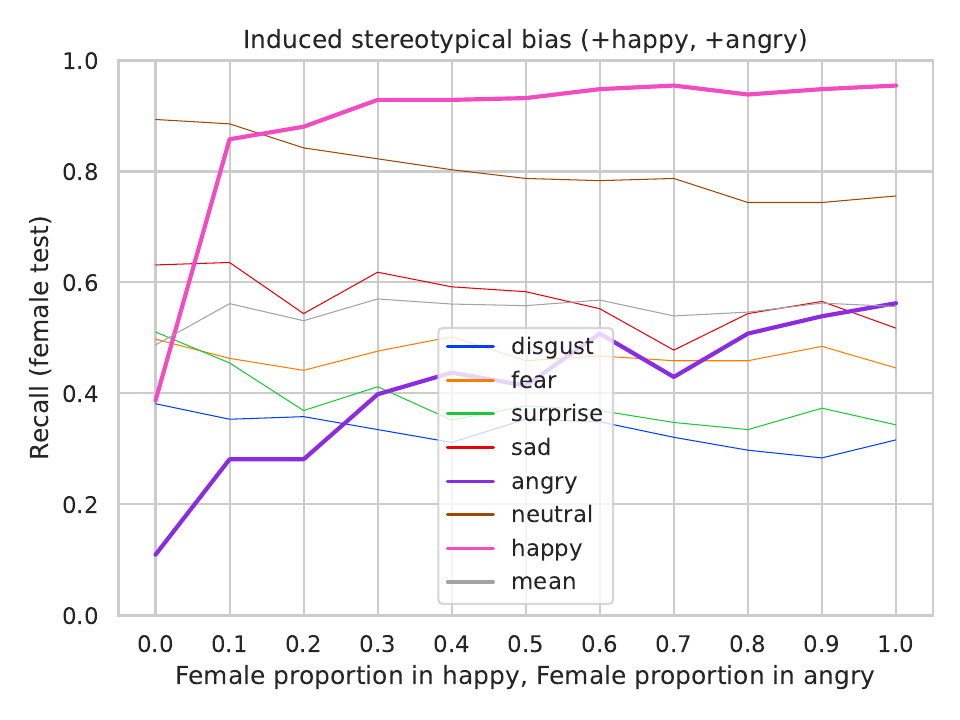}
        \label{fig:exp4b}}
        \subfigure[]{\includegraphics[width=0.32\linewidth]{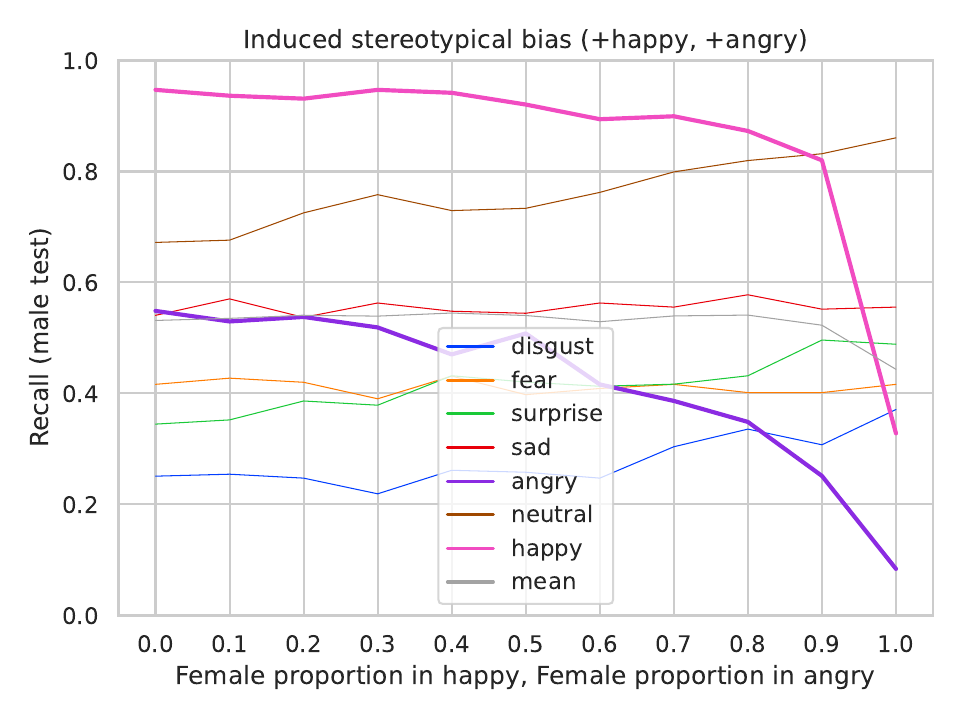}
        \label{fig:exp4c}}
        \caption{(a) Bias measures of each dataset stereotypically biased in happy and angry classes (same direction). (b) Recall per class for the female group. (c) Recall per class for the male group. For all three, in the horizontal axis, amount of induced bias.}
        \label{fig:exp4}
    \end{figure*}

    \textbf{Dataset bias metrics.} The results of studying the biased datasets with the proposed metrics are represented in Figure~\ref{fig:exp4a}. The ENS and SEI metrics (blue and orange lines, respectively) show stronger representational bias and unevenness compared to Experiments 2 and 3, and are consistent with Table~\ref{tab:exp4}, where the induced stereotypical bias also modifies the global number of samples for each gender group. The $\phi_C$ metric for stereotypical bias (green line), as expected, shows a strong stereotypical bias for ratios close to $0.0$ and $1.0$, and no stereotypical bias for the reference proportion $0.5$. Compared to Experiment 3, the stereotypical bias is weaker, slightly under $0.6$ in both cases, showing that the induced stereotypical bias is even larger than the one induced in Experiment 2, where only the happy class was perturbed.. Although the number of biased samples is the same in both experiments, the way we are biasing in this experiment means that a sample being of a specific gender gives us less information about which class it belongs to.

    \textbf{Model performance.} Recalls per class of the female (Figure~\ref{fig:exp4b}) and male groups (Figure~\ref{fig:exp4c}) show a situation similar to that of Experiment 3. The happy class shows a steep decline in the cases of extreme bias for both groups (proportion $0.0$ for female and $1.0$ for male) and a weaker decline throughout the range. The recall of the angry class shows the same behavior relative to the female proportion in the class as in Experiment 3, which in this experiment is increasing with the induced bias parameter for the female group and decreasing with the parameter for the male group. In both groups, the relationship for angry is more linear than in the happy class, although with a more steep variation when removing examples (proportion $<0.5$ for the female group and $>0.5$ for the male group) than when adding them. The recall of the rest of the classes shows a relatively weak variation (under $0.2$). Interestingly, the recall variations under for the rest of the classes seem to be coherent and evolve in an inverse way compared to those of the happy and angry classes.

    \textbf{Conclusions.} When the induced stereotypical bias affects multiple classes, with the same gender group underrepresented in all of them, the effect of the bias can compound. Although the effect for the individual perturbed classes does not significantly vary from biasing each single class, in this scenario the bias can start affecting the unbiased classes. Generally, unbiased classes show a recall variation inverse to that of biased classes. That is, if a gender group is sufficiently underrepresented in the stereotypically biased classes, the model will start to over-predict the other unbiased classes, increasing their recall.

    \subsection{Overall conclusions}

    From these experiments, we have seen how, for the FER problem, stereotypical bias on the gender component has an overall bigger impact than representational bias. In this case, representational bias has a negligible impact and only in extreme cases does it worsen the predictions of the minority group. We have also observed that stereotypical bias behavior changes significantly between classes, such as ``happy'' and ``angry''. While the easier ``happy'' class seems to degrade only in extreme cases, the ``angry'' class, harder to recognize by the models, degrades under any amount of bias. Although we have not found significant interactions between stereotypically biased classes, the results show that when multiple stereotypical biases affect several classes in the same direction, they can induce bias in the rest of the classes, even if those classes are balanced. 
    
    It is important to note that these results are only valid for this specific problem and context, and centered on one demographic component only. Despite this, they already show how different types of measurable bias affect the model in particular ways. We expect different problems and models to be affected in different ways, but, nevertheless, having better tools to assess and quantify dataset bias can be useful when studying those specific situations and how to mitigate those biases.

\def\datasets{
{adfes/$\text{ADFES}~^\text{LAB}$},
{affectnet/$\text{AFFECTNET}~^\text{ITW-I}$},
{caer-s/$\text{CAER-S}~^\text{ITW-M}$},
{ck/$\text{CK}~^\text{LAB}$},
{ck+/$\text{CK+}~^\text{LAB}$},
{fer2013/$\text{FER2013}~^\text{ITW-I}$},
{ferplus/$\text{FER+}~^\text{ITW-I}$},
{expw/$\text{EXPW}~^\text{ITW-I}$},
{gemep/$\text{GEMEP}~^\text{LAB}$},
{isafe/$\text{iSAFE}~^\text{LAB}$},
{kdef/$\text{KDEF}~^\text{LAB}$},
{mmafedb/$\text{MMAFEDB}~^\text{ITW-I}$},
{nhfier/$\text{NHFIER}~^\text{ITW-I}$},
{oulu-casia/$\text{Oulu-CASIA}~^\text{LAB}$},
{raf-db2/$\text{RAF-DB}~^\text{ITW-I}$},
{sfew/$\text{SFEW}~^\text{ITW-M}$},
{jaffe/$\text{JAFFE}~^\text{LAB}$},
{mug/$\text{MUG}~^\text{LAB}$},
{wsefep/$\text{WSEFEP}~^\text{LAB}$},
{liris-cse/$\text{LIRIS-CSE}~^\text{LAB}$}}

\FloatBarrier

\section{Local stereotypical bias results}\label{apA}

In Figures~\ref{figure:stereolocal_adfes} to \ref{figure:stereolocal_liris-cse} we provide the full contingency tables resulting from the demographic analysis of the twenty FER datasets, together with the NPMI and Z matrices for local stereotypical bias detection. The global stereotypical bias and representational bias results of these datasets are available in the main paper.

\foreach \fname/\dataset in \datasets
{\begin{figure*}[!ht]
    \graphicspath{{images/appendix_stereolocal/}}
    \centering
    \resizebox{\textwidth}{!}{
        \includegraphics{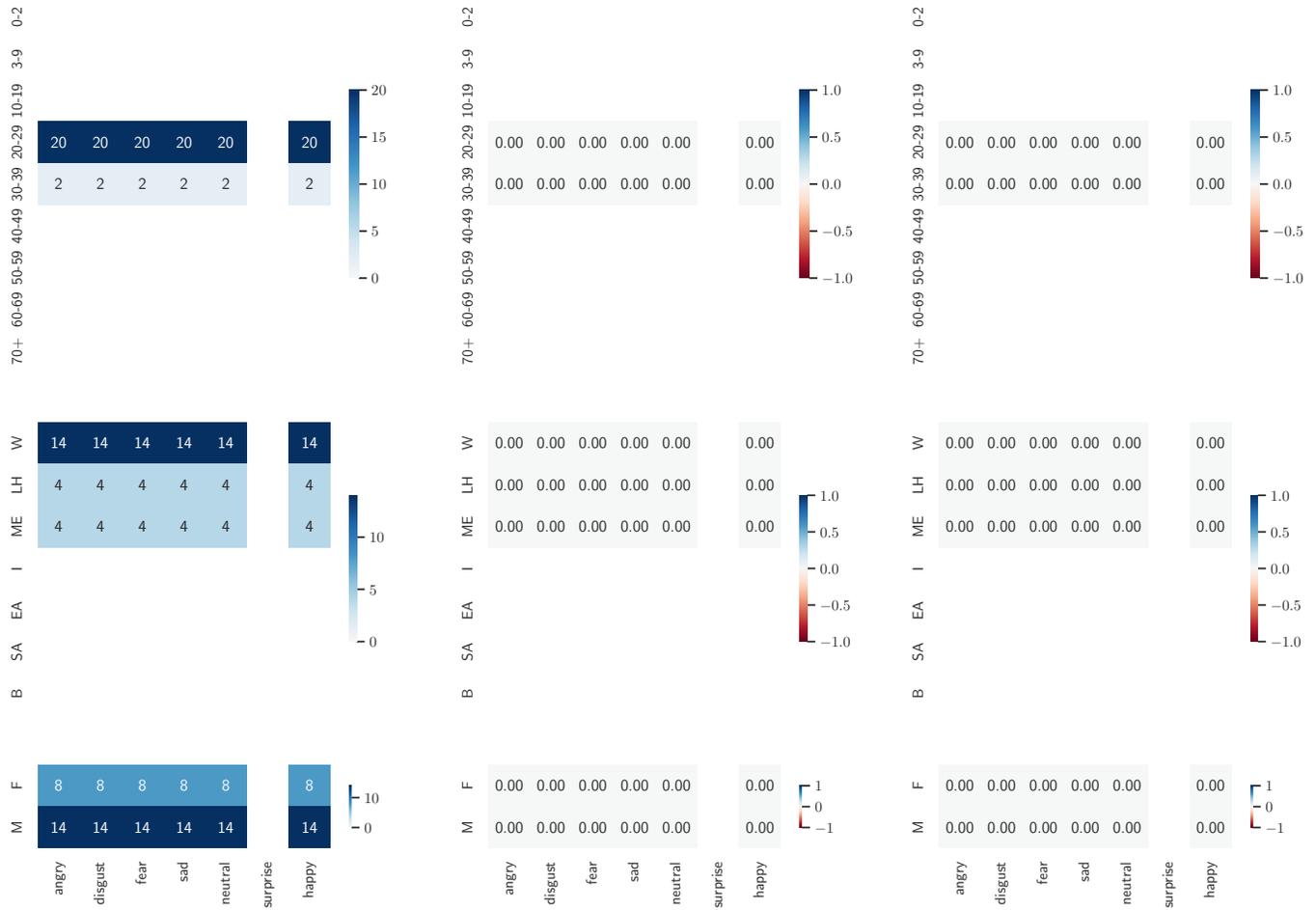}}
    \caption{Local stereotypical bias metrics for the \dataset \:dataset across the three demographic components. The first column corresponds to the raw number of samples in each combination of demographic group and target label, the second column corresponds to the NPMI metric and the third column to the Z metric. For the two metrics, higher absolute values correspond to higher local stereotypical bias. Negative values correspond to underrepresentation, and positive values to overrepresentation.}
    \label{figure:stereolocal_\fname}
\end{figure*}}

\end{appendices}

\FloatBarrier
\clearpage
\bibliographystyle{IEEEtran}
\bibliography{bstcontrol,FER_bibtex,reviewers}

\begin{IEEEbiography}[{\includegraphics[width=1in,height=1.25in,clip,keepaspectratio]{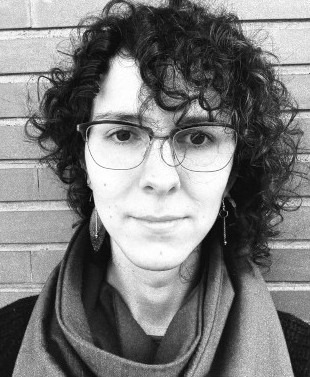}}]{Iris Dominguez-Catena} 
received the B.Sc. and M.Sc. degrees in computer science from the Public University of Navarra, in 2015 and 2020, respectively. She worked for several private software companies from 2015 to 2018. She is currently a Ph.D. candidate with the Public University of Navarra, working on demographic bias issues in Artificial Intelligence. Her research interests focus on AI fairness, bias detection and mitigation, and other ethical problems of AI deployment in society.
\end{IEEEbiography}

\begin{IEEEbiography}[{\includegraphics[width=1in,height=1.25in, clip, keepaspectratio]{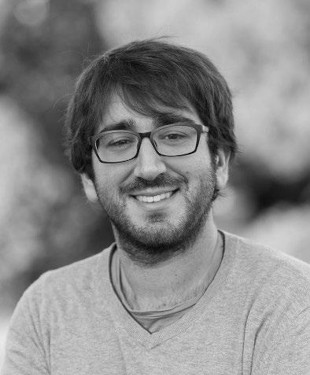}}]{Daniel Paternain}
received the M.Sc. and Ph.D. degrees in computer science from the Public University of Navarra, Pamplona, Spain, in 2008 and 2013, respectively. He is currently Associate Professor with the Department of Statistics, Computer Science and Mathematics. He is also the author or coauthor of almost 40 articles in journals from JCR and more than 50 international conference communications. His research interests include both theoretical and applied aspects of information fusion, computer vision and machine learning.
\end{IEEEbiography}

\begin{IEEEbiography}[{\includegraphics[width=1in,height=1.25in,clip,keepaspectratio]{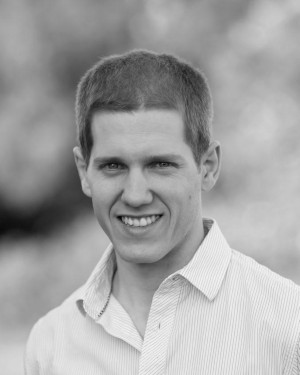}}]{Mikel Galar} (M14)
received the M.Sc. and Ph.D. degrees in computer science from the Public University of Navarra, Pamplona, Spain, in 2009 and 2012, respectively. He is currently an Associate Professor at the Public University of Navarra. He is the author of 50 published original articles in international journals and more than 80 contributions to conferences. He is a co-author of a book on imbalanced datasets and a book on large-scale data analytics. His research interests are  machine learning, deep learning, ensemble learning and big data. He  received the extraordinary prize for his PhD thesis from the Public University of Navarre and the 2013 IEEE Transactions on Fuzzy System Outstanding Paper Award (bestowed in 2016). 
\end{IEEEbiography}

\vfill

\end{document}